\documentclass[10pt,conference]{IEEEtran}

\usepackage{etoolbox}
\makeatletter
\def\ps@headings{%
\def\@oddhead{\mbox{}\scriptsize\rightmark \hfil \thepage}%
\def\@evenhead{\scriptsize\thepage \hfil \leftmark\mbox{}}%
\def\@oddfoot{}%
\def\@evenfoot{}}
\makeatother
\usepackage{xcolor}
\usepackage{amsfonts}
\usepackage{mathrsfs}
\usepackage{amsfonts}
\usepackage{amssymb}
\usepackage{graphicx}
\usepackage{epsfig}
\usepackage{epstopdf}
\usepackage{psfrag}
\usepackage{amsmath}
\usepackage{array}
\usepackage{cite,graphicx,amsmath,amssymb,color}
\usepackage{algorithmic}
\usepackage{algorithm}
\usepackage{booktabs} 
\usepackage{tikz}
\usepackage{amsmath}

\usepackage{algorithm}
\usepackage{algorithmic}
\usepackage{stmaryrd}
\usepackage{multirow}
\usepackage{subfig}
\usepackage{graphicx,times,amsmath} 

\usepackage{url}
\usepackage{enumerate}

\IEEEoverridecommandlockouts

\begin{document}
\bibliographystyle{IEEEtran}

\title{A Multi-Stage Optimization Framework for Deploying Learned Image Compression on FPGAs}


\author{
\IEEEauthorblockN{Jiaxun Fang} 
\IEEEauthorblockA{Shanghai Jiao Tong University, China\\ Email: oxygenfunction@sjtu.edu.cn} 
\and 
\IEEEauthorblockN{Li Chen} 
\IEEEauthorblockA{Shanghai Jiao Tong University, China\\ Email: hilichen@sjtu.edu.cn} 
}

\maketitle

\begin{abstract}

Deep learning-based image compression (LIC) has achieved state-of-the-art rate-distortion (RD) performance, yet deploying these models on resource-constrained FPGAs remains a major challenge. This work presents a complete, multi-stage optimization framework to bridge the gap between high-performance floating-point models and efficient, hardware-friendly integer-based implementations. First, we address the fundamental problem of quantization-induced performance degradation. We propose a Dynamic Range-Aware Quantization (DRAQ) method that uses statistically-calibrated activation clipping and a novel weight regularization scheme to counteract the effects of extreme data outliers and large dynamic ranges, successfully creating a high-fidelity 8-bit integer model. Second, building on this robust foundation, we introduce two hardware-aware optimization techniques tailored for FPGAs. A progressive mixed-precision search algorithm exploits FPGA flexibility to assign optimal, non-uniform bit-widths to each layer, minimizing complexity while preserving performance. Concurrently, a channel pruning method, adapted to work with the Generalized Divisive Normalization (GDN) layers common in LIC, removes model redundancy by eliminating inactive channels. Our comprehensive experiments show that the foundational DRAQ method reduces the BD-rate overhead of a GDN-based model from $30\%$ to $6.3\%$. The subsequent hardware-aware optimizations further reduce computational complexity by over $20\%$ with negligible impact on RD performance, yielding a final model that is both state-of-the-art in efficiency and superior in quality to existing FPGA-based LIC implementations.

\end{abstract}


\section{Introduction}
The explosive growth of visual data has driven significant demand for more efficient image compression technologies. In recent years, deep learning-based image compression (LIC) has emerged as a powerful paradigm, with end-to-end trained models consistently surpassing traditional codecs like JPEG2000 and HEVC in rate-distortion (RD) performance. These models replace hand-crafted modules with powerful nonlinear transforms learned from data, enabling superior feature representation and compression.

Despite their success, the practical deployment of LIC models, particularly on resource-constrained edge devices such as FPGAs, is severely hampered by their high computational complexity and memory footprint. These models typically rely on 32-bit floating-point (FP32) arithmetic and contain millions of parameters, making them unsuitable for low-power, real-time applications. Model quantization, the process of converting FP32 parameters and activations to low-bit-width integers (e.g., INT8), is an essential technique for enabling hardware acceleration. However, the direct application of standard quantization methods, such as those successful in image classification, leads to a catastrophic drop in RD performance for LIC models.

This paper addresses this challenge through a comprehensive, two-stage optimization framework. Our primary thesis is that achieving high-performance, hardware-friendly LIC requires tackling the problem at two distinct levels: first, solving the fundamental data distribution challenges that cause quantization failure, and second, applying advanced, hardware-aware optimizations to the resulting integer model to maximize efficiency.

\begin{figure*}
    \centering
    \includegraphics[width=0.95\linewidth]{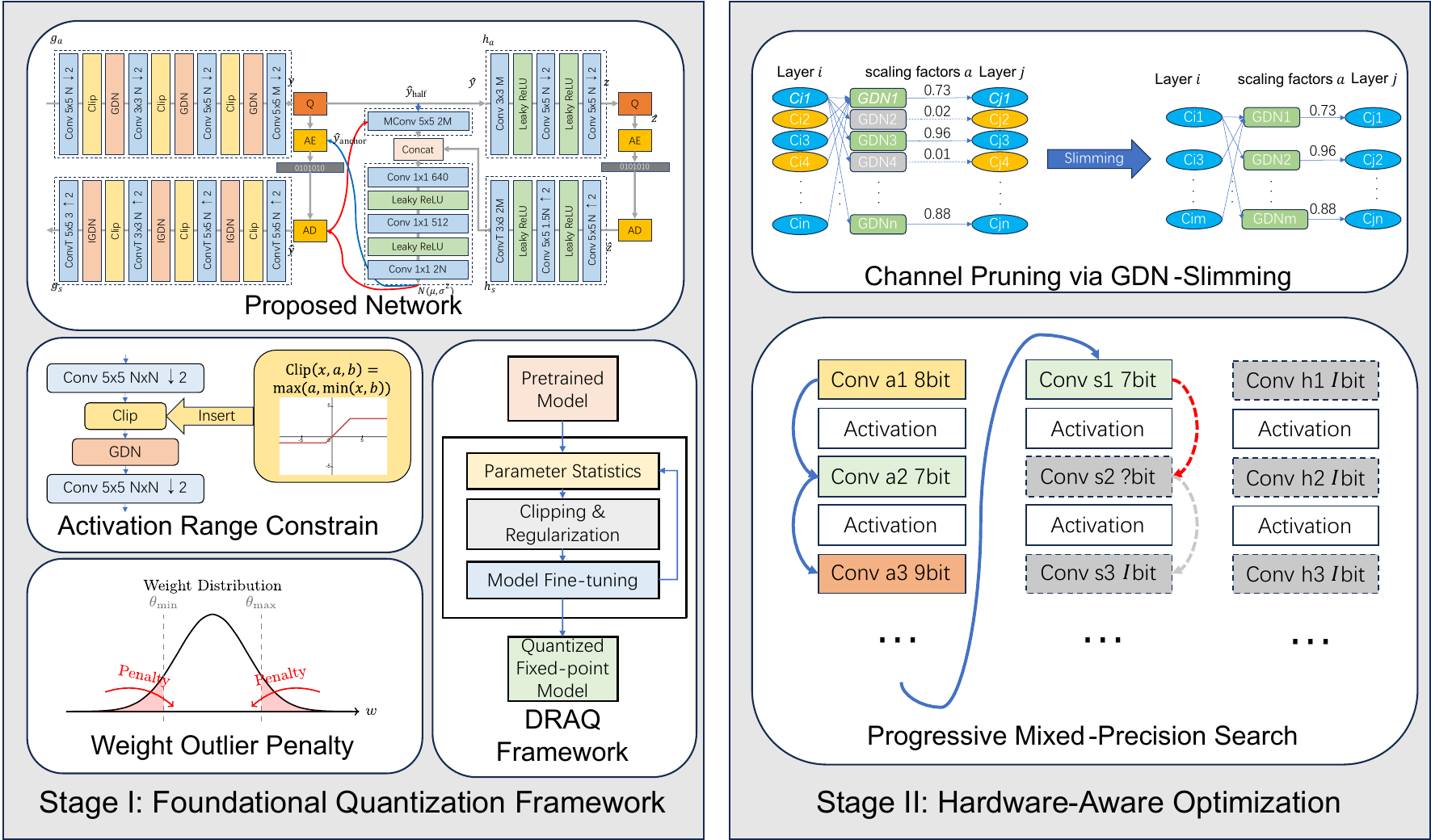}
    \caption{An overview of our proposed two-stage optimization framework. \textbf{Stage I}, the Dynamic Range-Aware Quantization (DRAQ) framework, addresses fundamental data distribution challenges by constraining activation ranges and penalizing weight outliers to create a high-fidelity integer model. \textbf{Stage II} applies hardware-aware optimizations tailored for FPGAs, including Channel Pruning via GDN-Slimming and a Progressive Mixed-Precision Search, to systematically balance performance and hardware complexity.}
    \label{fig:two_stage}
\end{figure*}

The first part of our work investigates the root causes of this performance degradation. We identify that the data distributions within LIC models are fundamentally different from those in typical classification networks. They are characterized by \textbf{(1) extreme outliers} in weight distributions that poison the quantization range, and \textbf{(2) large, channel-variant dynamic ranges} in activations that lead to significant information loss. To solve this, we propose a \textbf{Dynamic Range-Aware Quantization (DRAQ)} framework, as illustrated in Stage I of Fig.~\ref{fig:two_stage}.. This foundational method uses statistically-calibrated activation clipping and a novel weight regularization scheme to produce a high-fidelity 8-bit integer model that closely matches the performance of its FP32 counterpart.

The second part of our work addresses the fact that even a high-quality INT8 model is not optimally efficient for FPGAs. FPGAs offer unique opportunities for co-design that fixed-function hardware like GPUs cannot. As shown in Stage II of Fig.~\ref{fig:two_stage}, we exploit this by introducing two hardware-aware optimization techniques:

\begin{enumerate}
    \item  \textbf{Progressive Mixed-Precision Search:} Recognizing that different layers have varying sensitivity to quantization and that FPGAs can implement arbitrary integer bit-widths, we develop a search algorithm to automatically assign an optimal, non-uniform bit-width to each layer. This allows for a fine-grained tradeoff between computational complexity and RD performance, a key advantage of FPGA platforms.
    \item  \textbf{Channel Pruning via GDN-Slimming:} We observe significant redundancy in LIC models, where many channels remain inactive across diverse inputs. To eliminate this, we adapt the well-known \textit{slimming} channel pruning technique to work with the Generalized Divisive Normalization (GDN) layers that are crucial for LIC performance but lack the Batch Normalization (BN) layers that slimming typically relies on.
\end{enumerate}

By integrating these techniques, we create a complete pipeline that transforms a high-performance FP32 LIC model into a pruned, mixed-precision integer model optimized for FPGA deployment. Our contributions are:

\begin{itemize}
    \item A detailed analysis of the data distribution challenges unique to LIC model quantization.
    \item A foundational \textbf{DRAQ framework} that enables high-fidelity 8-bit quantization of both ReLU and advanced GDN-based LIC models.
    \item A \textbf{progressive search algorithm} for mixed-precision quantization that leverages FPGA flexibility to minimize model complexity while maintaining performance near the FP32 level.
    \item A novel adaptation of \textbf{channel pruning (slimming)} for GDN-based architectures, effectively reducing model redundancy and computational load.
    \item A comprehensive experimental validation demonstrating that our final optimized model is more efficient and achieves a better RD performance than existing FPGA-based LIC implementations.
\end{itemize}

This paper is organized as follows. Section II details the unique data distribution challenges in LIC models. Section III presents our foundational DRAQ framework. Section IV describes the hardware-aware mixed-precision and pruning optimizations. Section V provides a comprehensive experimental evaluation, and Section VI concludes the paper. This work is an extension of the research detailed in.

\section{Related Works}

\subsection{Learned Image Compression}

\begin{figure}
    \centering
    \includegraphics[width=0.95\linewidth]{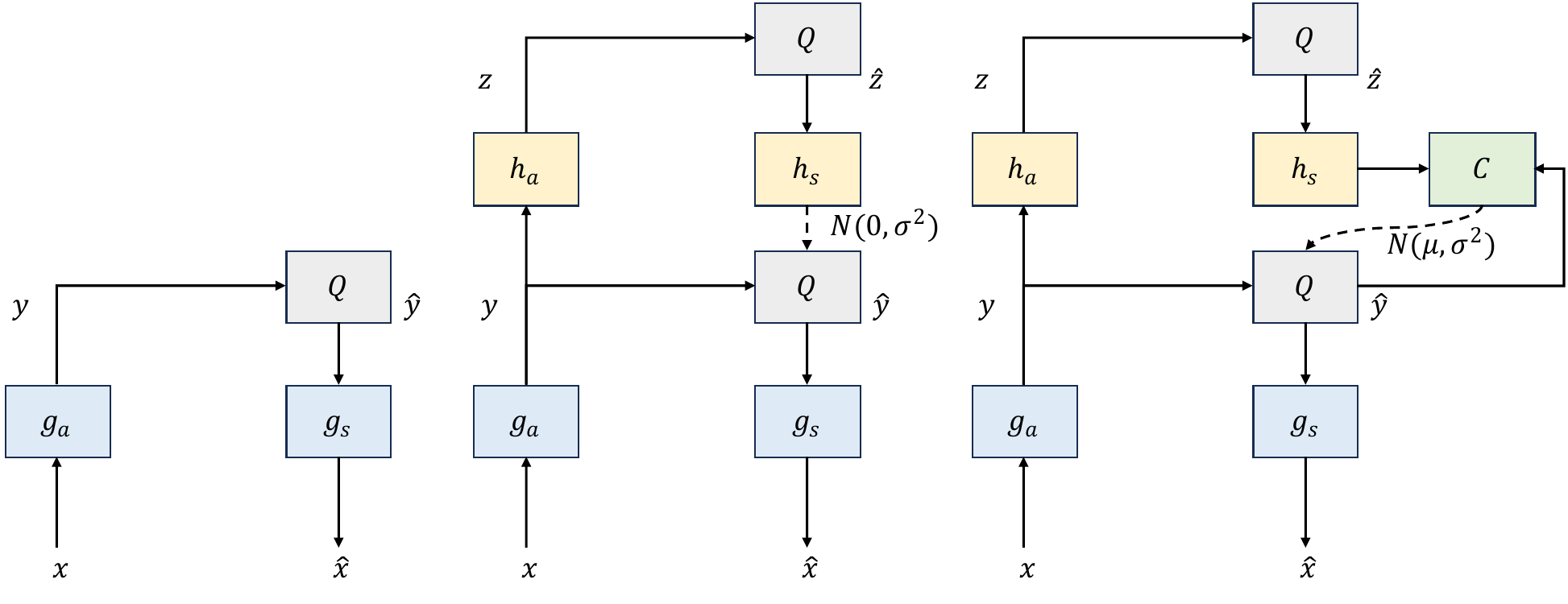}
    \caption{The evolution of seminal LIC architectures. (Left) The foundational variational autoencoder framework where latents $\mathbf{y}$ are quantized and reconstructed. (Center) The addition of a hyperprior network ($h_a, h_s$) that uses side information $\mathbf{z}$ to model the latent distribution, improving compression rate. (Right) The joint architecture, which further incorporates an autoregressive context model ($C$) to create a more powerful conditional probability model for the latents, leading to superior performance.}
    \label{fig:lic_evolution}
\end{figure}

The paradigm of image compression has been significantly influenced by deep learning, leading to end-to-end learned image compression (LIC) models that now outperform traditional codecs like VVC in rate-distortion (R-D) performance. The foundational architecture for most current LIC models is based on a variational autoencoder framework, first introduced by Ballé et al. \cite{balle2016end}. This framework consists of a nonlinear analysis transform $g_a$ that maps an image $ \mathbf{x} $ to a latent representation $ \mathbf{y} $, a uniform quantizer, and a nonlinear synthesis transform $g_s$ that reconstructs the image $ \hat{\mathbf{x}}$ from the quantized latents $\hat{\mathbf{y}}$ (as shown on the left in Fig.~\ref{fig:lic_evolution}). To enable end-to-end training via gradient descent, the non-differentiable quantization step is typically replaced by adding uniform noise during training. The models are optimized by minimizing a rate-distortion loss: $\mathcal{L}  =R+\lambda \cdot D $, where $R$ is the estimated bitrate from an entropy model and $D$ is the distortion between $ \mathbf{x} $  and $ \hat{\mathbf{x}}$, commonly measured by Mean Squared Error (MSE) or MS-SSIM.

A major breakthrough was the introduction of a hyperprior network \cite{balle2018variational}, which captures spatial redundancies within the latent representation $ \mathbf{y} $. The hyperprior acts as a side-information channel, generating parameters (e.g., the standard deviation $\sigma
$) for a conditional Gaussian probability model of the latents, thereby enabling more accurate entropy modeling and improved compression efficiency (center diagram of Fig.~\ref{fig:lic_evolution}). Building upon this, Minnen et al. \cite{minnen2018joint} incorporated a spatial context model, employing an autoregressive component that uses previously decoded latent elements to predict the distribution (both mean $\mu$ and standard deviation $\sigma
$) of the current element (right diagram of Fig.~\ref{fig:lic_evolution}). This joint autoregressive and hyperprior architecture became a seminal model, surpassing the performance of the BPG codec and setting the standard for subsequent research.

Recent advancements in LIC can be broadly categorized into two main streams: optimizing the network architecture of the transforms and refining the entropy models.

\subsubsection{Advancements in Network Architecture}
To enhance the expressive power of the nonlinear transforms, researchers have integrated more sophisticated neural network modules. Chen et al. \cite{chen2021end} introduced non-local attention mechanisms to capture global correlations, allowing the model to allocate bits adaptively based on content importance. To mitigate the high computational cost of attention, Cheng et al. \cite{cheng2020learned} demonstrated that stacked residual blocks could effectively expand the receptive field, achieving comparable performance with lower complexity. More recently, the success of Transformers in vision tasks has spurred their adoption in LIC. Architectures incorporating Swin Transformer blocks \cite{lu2021transformer} and hybrid CNN-Transformer designs \cite{liu2023learned} have been proposed to effectively model both local and global information, leading to state-of-the-art R-D performance.

\subsubsection{Advancements in Entropy Modeling}
Parallel to architectural improvements, significant efforts have been dedicated to designing more powerful and efficient entropy models. To better capture the complex, non-Gaussian distributions of latent features, Cheng et al. \cite{cheng2020learned} proposed using Gaussian Mixture Models (GMMs) instead of a single Gaussian, improving probability estimation accuracy. The context model has also been a focus of optimization. To address the inherent serial processing bottleneck of spatial autoregression, He et al. proposed a parallelizable checkerboard context model \cite{he2021checkerboard}, which divides latents into two groups that can be processed in two parallel steps, drastically accelerating decoding with minimal performance loss. They later introduced a channel-wise autoregressive model \cite{he2022elic}, which groups channels and performs autoregression across channel groups, striking a balance between performance and computational complexity.

\subsection{Model Quantization for Efficient Inference}

Deploying large neural networks, including LIC models, on resource-constrained devices necessitates model compression. Quantization is a key technique that reduces model size, memory bandwidth, and computational energy by converting high-precision floating-point parameters and activations (e.g., FP32) to low-bit integers (e.g., INT8). The two dominant quantization paradigms are Post-Training Quantization (PTQ) and Quantization-Aware Training (QAT).

\subsubsection{Post-Training Quantization (PTQ)}
PTQ quantizes a pre-trained model with minimal or no retraining, making it a fast and data-efficient solution. The core challenge in PTQ is determining the optimal clipping range (quantization parameters) for each tensor to balance clipping and rounding errors. While simple methods like Min-Max calibration are sensitive to outliers, others minimize metrics like Mean Squared Error (MSE) between the original and quantized tensors. More advanced techniques have been developed to improve PTQ performance. For instance, Cross-Layer Equalization (CLE) \cite{nagel2019data} rectifies large variations in weight ranges across consecutive layers by rescaling them, making the model more robust to quantization. AdaRound \cite{nagel2020up} is another powerful method that learns a task-aware rounding scheme for weight quantization, often outperforming standard nearest-rounding and preserving accuracy at lower bit-widths.

\subsubsection{Quantization-Aware Training (QAT)}
QAT simulates the effect of quantization during the training or fine-tuning process, allowing the model to adapt its weights to the quantization error. This is achieved by inserting \textit{fake}  quantization nodes into the network graph, which quantize and de-quantize tensors during the forward pass. Since the rounding function has zero or undefined gradients, the Straight-Through Estimator (STE) \cite{bengio2013estimating} is ubiquitously used to approximate the gradient of the quantizer, enabling end-to-end training. QAT generally yields higher accuracy than PTQ, especially for aggressive quantization (e.g., below 8 bits), but requires access to the training dataset and significantly more computational resources for retraining.

Beyond quantization, model pruning, which removes redundant weights or structures, is often used as a complementary compression technique. Methods that combine structured pruning or neural architecture search (NAS) with quantization have shown promise in automatically discovering highly efficient and accurate compressed models \cite{wang2020apq}.

\section{The Challenge in Quantization of LIC Models}

\subsection{Data Distributions}

Standard quantization-aware training (QAT) fails for LIC models primarily because it assumes data distributions that do not hold true. Our analysis revealed two fundamental problems.

\subsubsection{Extreme Weight Outliers}
\begin{figure*}
    \centering
    \includegraphics[width=0.9\linewidth]{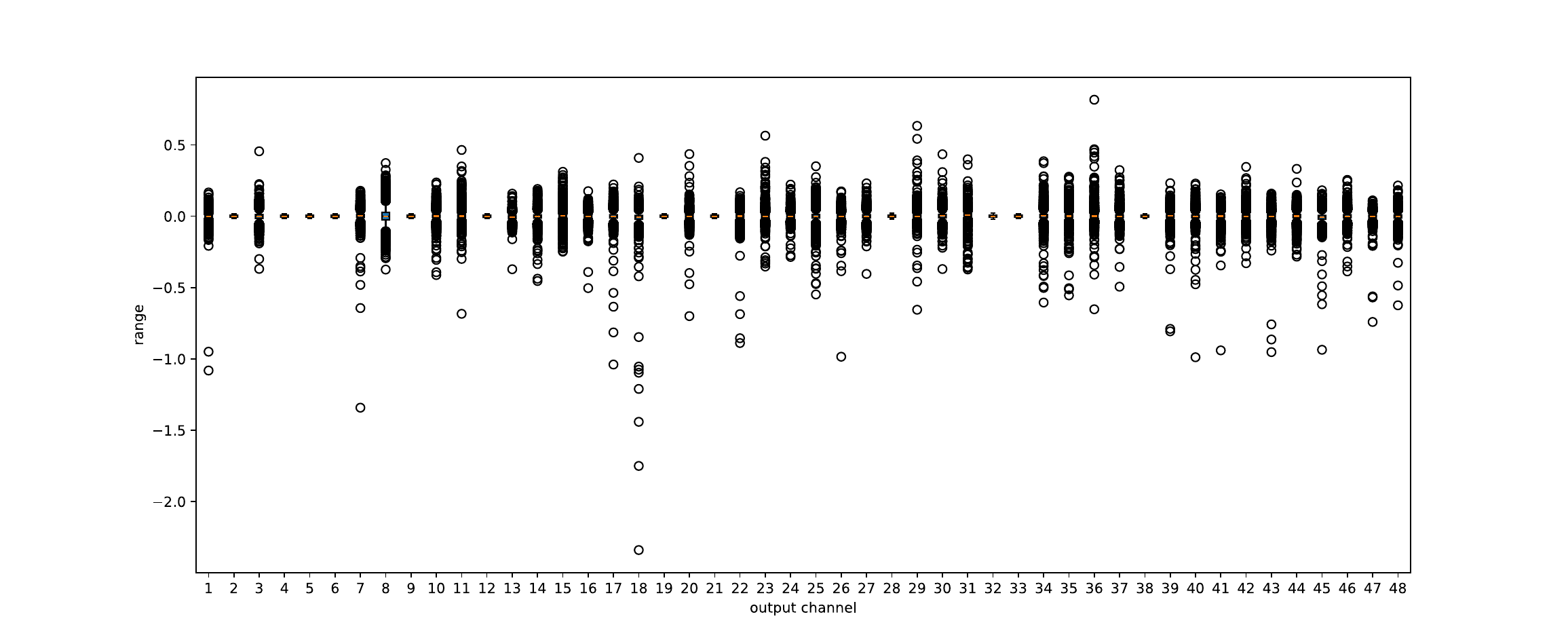}
    \caption{The challenge of quantizing LIC models, illustrated by the weight distribution of a single representative convolutional layer. The boxplot for each output channel reveals a high concentration of weights near zero (indicated by the small boxes), but also a tiny fraction of extreme outliers (circles) that are orders of magnitude larger. These outliers dictate an inefficiently wide quantization range, degrading precision for the majority of weights and motivating our proposed outlier-aware framework.}
    \label{fig:weight_dis}
\end{figure*}

A primary obstacle to quantizing LIC models lies in the unique statistical properties of their network weights. Our analysis reveals a consistent and challenging data distribution across all layers, which is fundamentally different from that found in standard classification networks. As illustrated by a representative convolutional layer in Fig.~\ref{fig:weight_dis}, the weights exhibit two defining characteristics:

\begin{itemize}
    \item \textbf{High Concentration:} The vast majority of weights are densely clustered in a very narrow range around zero, indicated by a small interquartile range (IQR).
    \item \textbf{Extreme Outliers:} This dense cluster is accompanied by a tiny fraction of extreme outliers (e.g., $\le0.01\%$), with magnitudes that can be orders of magnitude larger than the core distribution.
\end{itemize}

These properties, particularly the presence of extreme outliers, render conventional quantization strategies like linear min-max quantization highly ineffective. The outliers dictate an excessively wide quantization range, which leads to two critical problems:

\begin{enumerate}
    \item Significant Information Loss: The vast majority of concentrated weights are aggressively mapped to only a few quantization levels near zero. This erases the subtle but crucial variations among these weights, crippling the model's expressive power and leading to severe performance degradation.
    \item Quantization Inefficiency: A large portion of the available quantization bins remains unused, as they correspond to the vast empty space between the core distribution and the extreme outliers. This represents a significant waste of bit-width resources.
\end{enumerate}

For instance, in the distribution shown in Fig.~\ref{fig:weight_dis}, a single outlier near -2.5 dictates the quantization range. Removing a handful of such outliers ($<0.001\%$ of the total) could halve the data range, effectively doubling the quantization precision for the remaining weights. Therefore, we identify these rare but extreme weight outliers as the principal barrier to achieving efficient, high-fidelity quantization in LIC models. This analysis motivates the need for a quantization framework that is robust to such distributions.

\subsubsection{Activation Dynamic Range Variance}
\begin{figure}
    \centering
    \subfloat[Maximum (red) and minimum (blue) activation values per channel.]{
        \label{fig:act_minmax}
        \includegraphics[width=0.95\linewidth]{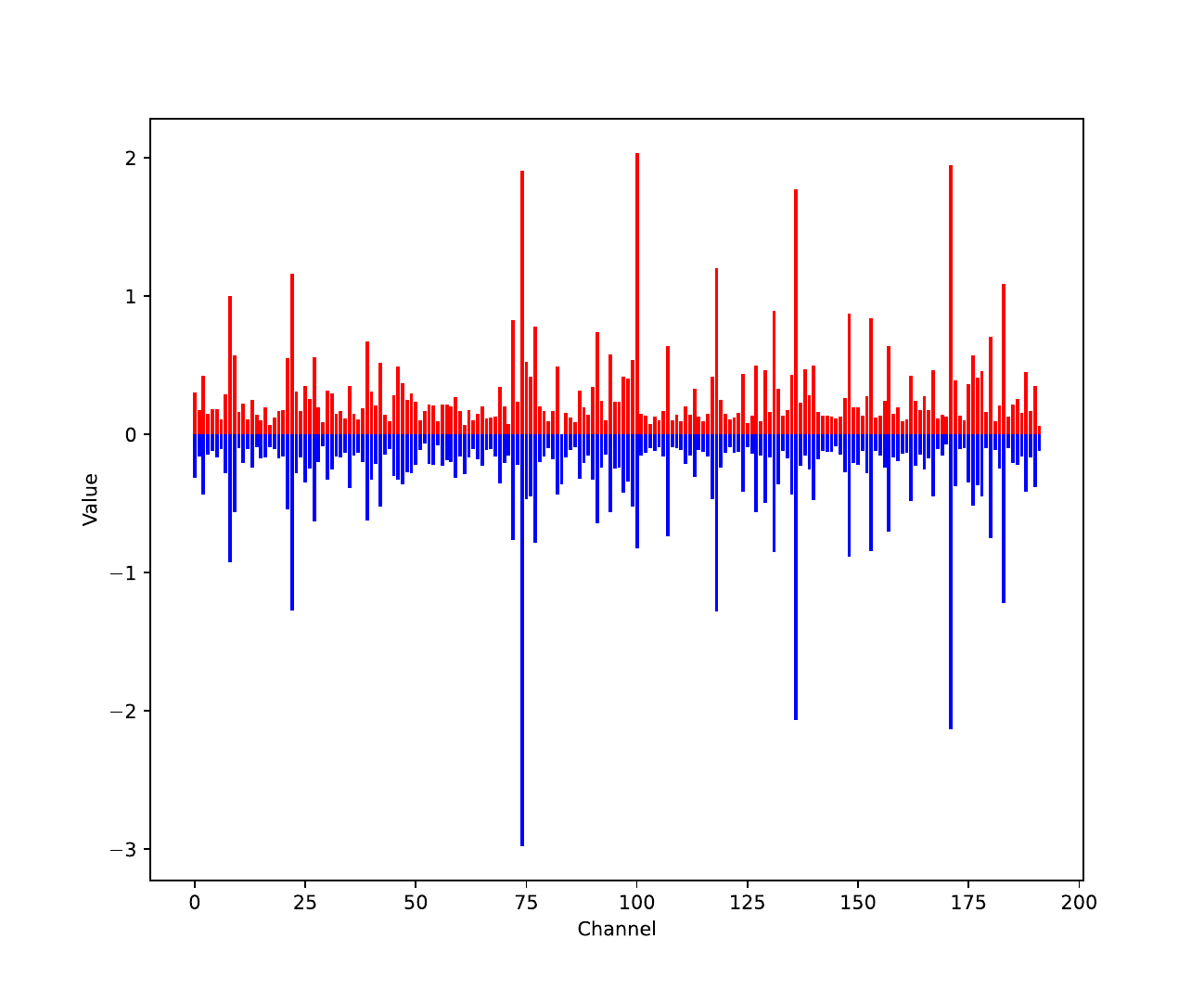}
    }
    \\
    \subfloat[Frequency histograms of select channels, with inset showing detail near zero.]{
        \label{fig:act_frequnce}
        \includegraphics[width=0.95\linewidth]{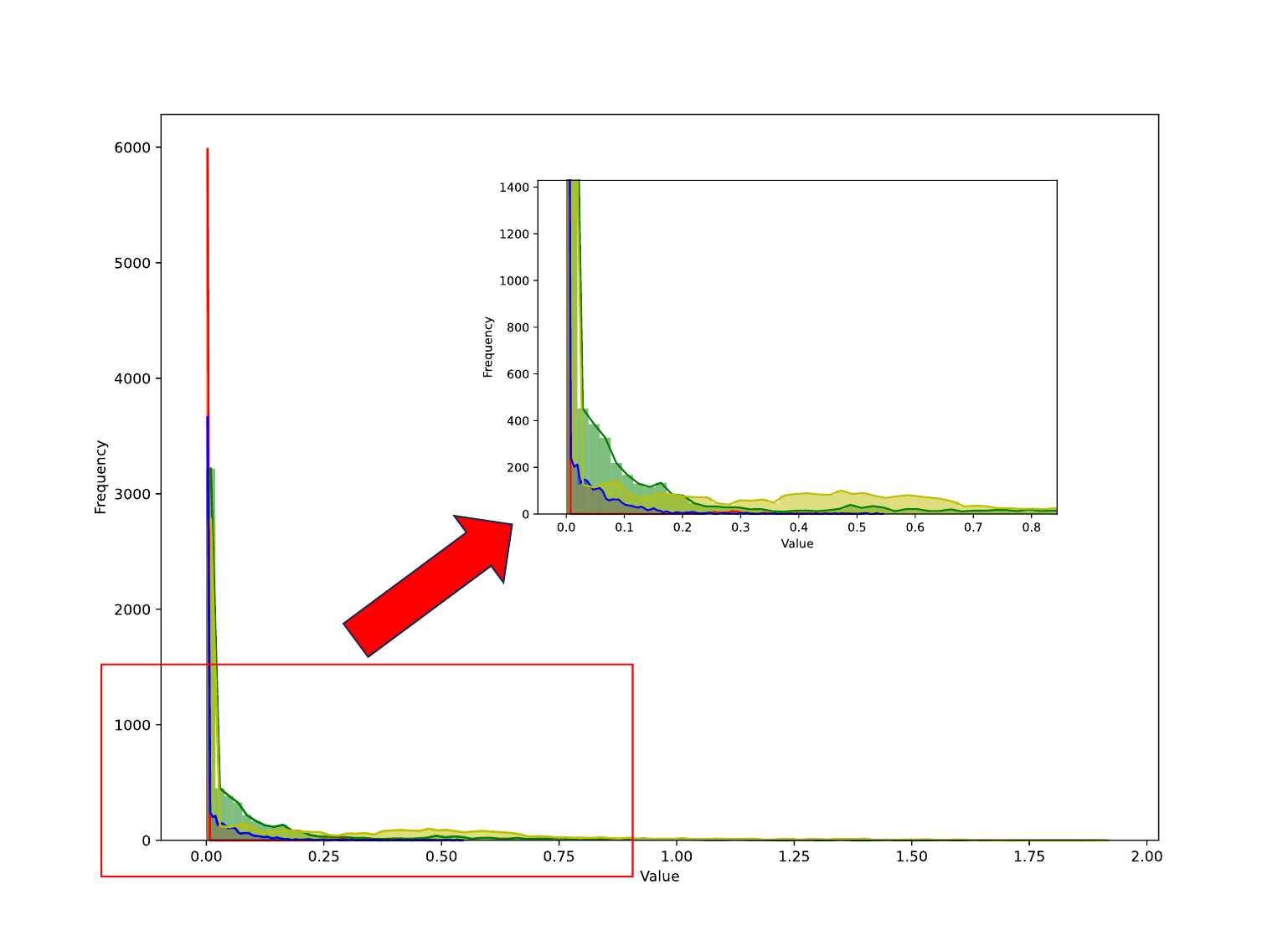}
    }
    \caption{The challenge of activation quantization, illustrated on a single representative layer. (a) The dynamic range of activations varies dramatically across different channels, with a few channels having ranges far exceeding the majority. (b) Histograms reveal that channels also possess distinct distributions. This high channel-wise heterogeneity makes a single, shared per-layer quantization scale highly suboptimal, leading to significant information loss for most channels.}
    \label{fig:act_dis}
\end{figure}

A second, equally critical challenge arises from the distribution of activation values. Unlike weights, the primary issue with activations is not isolated outliers but a significant channel-wise variance in their dynamic range. As shown in Fig. \ref{fig:act_dis}, activations from different channels within the same layer can exhibit vastly different statistical properties.

This heterogeneity becomes a major bottleneck for standard per-layer quantization, where a single scaling factor is applied to all channels in a layer. This shared factor is inevitably dictated by the channel with the widest dynamic range. Consequently, this leads to two detrimental effects:

\begin{enumerate}
    \item For low-range channels: The oversized scaling factor causes a catastrophic quantization error. Their entire data distribution, which may contain rich information, collapses into a few quantization bins near zero, effectively erasing their contribution. As seen in Fig. \ref{fig:act_dis}(a), many channels have a maximum activation value less than 5, while a few exceed 20. A uniform scale set by the latter would decimate the former.
    \item For high-range channels: Even within the channels that dictate the scaling factor, the activation values are often not uniformly distributed. Instead, they are typically concentrated in a sub-region of their total range (Fig. \ref{fig:act_dis}(b)), leading to the same issue of inefficient quantization bin utilization seen in weights.
\end{enumerate}

Therefore, applying a naive per-layer quantization scheme to activations results in a suboptimal trade-off, where precision is sacrificed for some channels while bit-width is wasted for others.While per-channel quantization can solve this, it introduces significant computational overhead, especially on hardware not optimized for it.

\subsubsection{Summary of Data Distribution Challenges}

In summary, our analysis identifies two fundamental obstacles to the effective quantization of learned image compression models:

\begin{enumerate}
    \item \textbf{Extreme Weight Outliers:} These force an overly large quantization range, which degrades precision for the vast majority of concentrated weights.
    \item \textbf{Large Channel-wise Dynamic Range Variance in Activations:} This makes a single per-layer scaling factor suboptimal, causing significant information loss in many channels.
\end{enumerate}

Any successful quantization framework for LIC must directly address these two data distribution challenges. This motivates our proposed approach, which is designed to be robust to both outlier-prone weights and heterogeneous activation ranges.

\subsection{The Quantization Challenge of GDN-based Architectures}
While Generalized Divisive Normalization (GDN) layers are critical for achieving better rate-distortion performance in full-precision LIC models \cite{balle2016end}, they represent a significant roadblock to effective quantization. Consequently, much of the existing work on LIC quantization sidesteps this issue by reverting to simpler, lower-performing ReLU-based backbones. Our analysis reveals that successfully quantizing GDN-based models requires overcoming two fundamental and interconnected challenges: exacerbated dynamic range and severe training instability.

\subsubsection{Exacerbated Dynamic Range}
The dynamic range problem, already present in ReLU models, is significantly magnified in GDN-based architectures. The activation values at the output of both convolutional and GDN layers exhibit extremely large, channel-variant dynamic ranges. Crucially, without the clipping effect of a ReLU function which discards negative values, the dynamic range of activations in GDN models is substantially wider than in their ReLU counterparts. This directly translates to larger quantization step sizes, which causes more aggressive information loss and leads to a precipitous drop in model performance post-quantization.

\subsubsection{Training Instability and Convergence Failure}
The inherent complexity of the GDN transformation makes Quantization-Aware Training (QAT) notoriously difficult. The GDN operation is defined as:
\begin{equation}
z_i = \frac{x_i}{(\beta_i + \sum_{j} \gamma_{ij} |x_j|^{\alpha_{ij}})^{\varepsilon_i}}.
\end{equation}

During QAT, the input tensor $x_i$ must be quantized. However, this quantized input appears in both the numerator and the denominator, and its absolute value is taken within a non-linear function. This complex data dependency, combined with the use of STE to approximate gradients, creates a highly unstable training dynamic. As noted in prior work \cite{he2022post}, this often leads to a failure to converge, rendering standard QAT methods ineffective for these architectures.

Addressing these twin challenges—taming the extreme dynamic range while ensuring stable training convergence—is therefore imperative to unlock the potential of high-performance LIC models in low-bitwidth, hardware-constrained environments. This motivates our development of a specialized quantization framework tailored for the unique properties of GDN.

\section{DRAQ: A Foundation For High-Fidelity Quantization}

To overcome the dual challenges of extreme weight outliers and large activation range variance, we propose a Dynamic Range-Aware Quantization (DRAQ) framework. Instead of treating quantization as a post-processing step, DRAQ integrates distribution control directly into the training process. The framework is composed of two core components designed to reshape the data distributions for higher quantization efficiency: \textbf{(1) Statistically-Calibrated Activation Clipping} to control activation ranges, and \textbf{(2) Outlier-Aware Weight Regularization} to suppress extreme weight values.

\begin{figure}
    \centering
    \includegraphics[width=0.95\linewidth]{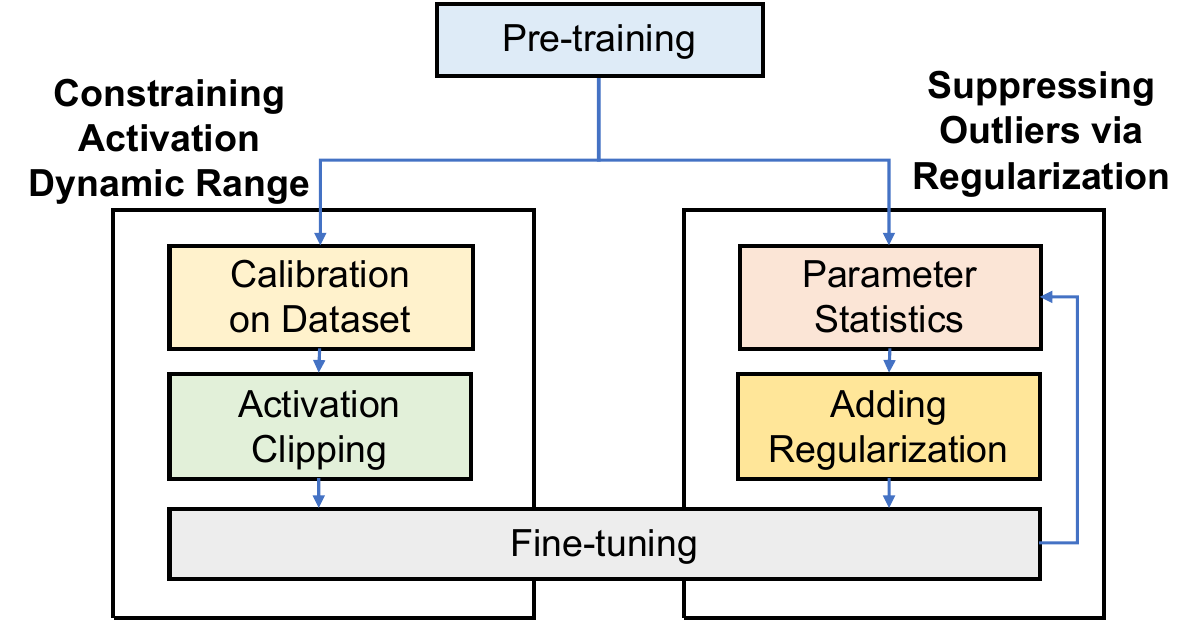}
    \caption{The workflow of our proposed Dynamic Range-Aware Quantization (DRAQ) framework. The process involves pre-training a baseline model, using a calibration set to determine optimal parameters for clipping and regularization, and then fine-tuning the model with these constraints to achieve a high-performance quantized model.}
    \label{fig:draq_flow}
\end{figure}

The overall workflow of our method is illustrated in Fig.~\ref{fig:draq_flow}. The process begins by training a high-quality baseline quantized model. This model is then used on a calibration dataset to gather detailed statistics about its weight and activation distributions. These statistics are used to determine the optimal parameters for our clipping and regularization functions. Finally, the model undergoes a constrained fine-tuning process, where the new components are integrated to produce the final, highly efficient quantized model. The specifics of each component are detailed below.

\subsection{Constraining Activation Dynamic Range}

To manage the large and variant activation ranges, we replace the standard ReLU activation function with a ClippedReLU, defined as:

\begin{equation}
    \text{ClippedReLU}(x, \theta) = \max(0, \min(x, \theta)) ,
\end{equation}

The key to our method is how we determine the upper-bound threshold $\theta$. Instead of using a fixed or naively trainable value, we determine it statistically for each layer $l$ from a calibration dataset. We compute the mean $\mu^l$ and standard deviation $\sigma^l$ of that layer's activation distribution and set the threshold $\theta^l$ as:

\begin{equation}
    \theta^l = \mu^l + k\sigma^l
\end{equation}

This approach, inspired by the three-sigma rule, ensures most of the activation distribution is preserved while clipping extreme tails. We further observed that the optimal value for $k$ is positively correlated with the rate-distortion tradeoff parameter $\lambda$ used in the loss function ($L = R + \lambda D$). Higher $\lambda$ values, which prioritize quality (PSNR) over rate (bpp), benefit from a larger $k$ that preserves more detail. We model this relationship with a simple but effective linear function:

\begin{equation}
    k = 625\lambda + 2
\end{equation}

This heuristic automatically adapts the clipping strength to the target bitrate, eliminating the need for manual hyperparameter tuning.

\subsection{Suppressing Weight Outliers via Regularization}

Directly clipping the extreme weight outliers is a blunt instrument that can cause irreversible information loss. We propose a more gentle approach by introducing a regularization term into the loss function during the fine-tuning stage to discourage, rather than forbid, extreme values.

First, we define outliers based on their statistical rarity. For each layer, we sort the weights from the pre-trained model and identify the values at the $\alpha$ and $1-\alpha$ percentiles (e.g., $\alpha=0.001$), which we denote as $\theta_{min}$ and $\theta_{max}$, respectively. We then introduce a regularization term to the standard rate-distortion loss function that penalizes any weight $w$ falling outside this range:

\begin{equation}
\mathcal{L}_{reg} = \beta \sum_{w>\theta_{max}} |w-\theta_{max}| + \beta \sum_{w<\theta_{min}} |w-\theta_{min}|,
\end{equation}

The total loss becomes  $\mathcal{L}  =R+\lambda \cdot D + \mathcal{L}_{reg}$, where $\beta$ is a coefficient controlling the regularization strength. This penalty acts as a \textit{soft} constraint, gently pulling outlier weights back towards the main distribution during fine-tuning without disrupting the learned features. The thresholds $\theta_{min}$ and $\theta_{max}$ are periodically re-calibrated during fine-tuning to adapt to the evolving weight distribution.

\subsection{Quantizing Generalized Divisive Normalization (GDN) Layers}
\subsubsection{Decomposing and Quantizing the GDN Operation}
For efficient hardware mapping and to simplify the complex GDN transformation, we adopt the common inference-time simplification, which sets the exponents to one:
\begin{equation}
z_i = \frac{x_i}{\beta_i + \sum_{j} \gamma_{ij} |x_j|}.
\end{equation}
Our key insight for quantization is that this operation can be decomposed into standard, hardware-friendly primitives. The denominator, $\beta_i + \sum_{j} \gamma_{ij} |x_j|$, is functionally equivalent to a depthwise convolution operation applied to the absolute value of the input, with $\gamma$ as the kernel and $\beta$ as the bias. The entire GDN layer can therefore be implemented as a depthwise convolution followed by an element-wise division.

This decomposition informs our quantization scheme:
\begin{itemize}
    \item Parameters ($\gamma,\beta$): The learnable parameters are quantized using a standard per-channel symmetric quantization scheme, identical to how we handle conventional convolutional layers.
    \item Input Activation ($x$): A critical design choice is how to quantize the input tensor $x$, which is used in both the numerator and the denominator. While quantizing $x$ and its absolute value $|x|$ separately with unique scaling factors offers theoretical precision benefits, we found this approach consistently fails to converge during QAT. The instability likely arises from compounding errors from the STE interacting with the absolute value function.
\end{itemize}

Therefore, we adopt a simpler, more robust scheme: $x$ is quantized once, and its quantized absolute value is used in the denominator's depthwise convolution. This approach significantly improves training stability with a negligible impact on final performance.

\subsubsection{Calibrated Clipping for Dynamic Range Control}
Quantizing the GDN operation alone is insufficient. As established, the unbounded dynamic range of its inputs remains the primary bottleneck, leading to massive quantization error that negates the performance benefits of GDN.

To resolve this, we introduce the second, crucial part of our strategy: inserting a calibrated, two-sided clipping layer to tame the activation range before it enters the GDN operation (as depicted in Fig. \ref{fig:two_stage}). The clipping function is defined as:

\begin{equation}
\text{Clip}(x, a, b) = \max(a, \min(x, b)),
\end{equation}
where $a$ and $b$ are the lower and upper clipping bounds.

Crucially, these bounds are not arbitrary but are determined by the same statistics-based method used throughout our framework. After pre-training, we use a calibration dataset to find the layer-wise mean $\mu$ and standard deviation $\sigma$ , setting the bounds to:
\begin{equation}
a = \mu - k\sigma \quad \text{and} \quad b = \mu + k\sigma.
\end{equation}

This two-sided, statistically-grounded clipping is the key to stabilizing the training of quantized GDN models. It constrains the problematic dynamic range, allowing the model to be fine-tuned effectively and finally surpass the performance of simpler, quantized ReLU-based models.

\section{Hardware-Aware Optimization For FPGAS}
The DRAQ framework provides a high-fidelity 8-bit model. However, for a target like an FPGA, which offers fine-grained control over hardware resources, we can achieve even greater efficiency. A uniform 8-bit data path may be overkill for some parts of the network and insufficient for others. We introduce two advanced optimization techniques that build upon the DRAQ-quantized model to create a solution specifically tailored for FPGAs.

\subsection{Progressive Mixed-Precision Search for Hardware-Efficient LIC}
\subsubsection{The Progressive Search Algorithm}

FPGAs, unlike GPUs, are not restricted to 8/16/32-bit arithmetic; they can implement custom-width datapaths (e.g., 7-bit, 10-bit), allowing for a precise tradeoff between resource usage and numerical precision. We exploit this with a progressive search algorithm to find the optimal bit-width for each layer.

\begin{figure}
    \centering
    \includegraphics[width=0.8\linewidth]{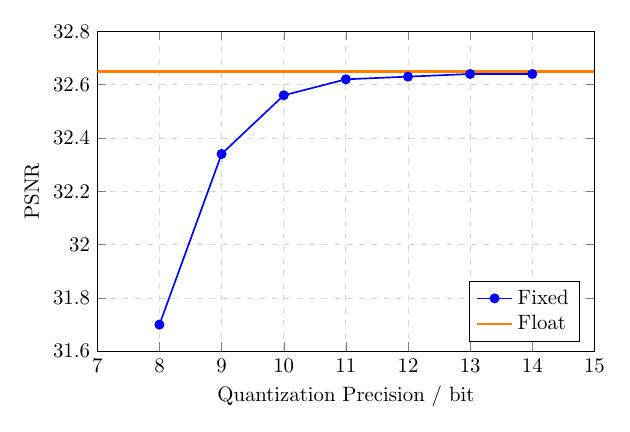}
    \caption{The law of diminishing returns in quantization precision. For a given RD trade-off point, performance (PSNR) saturates beyond a certain bit-width (the \textit{inflection point}, around 12 bits here). This motivates a mixed-precision approach to avoid wasting bits on layers that do not benefit from higher precision.}
    \label{fig:inflection_point}
\end{figure}

Our approach is motivated by the law of diminishing returns in quantization precision. As shown in Fig. \ref{fig:inflection_point}, model performance improves as bit-width increases, but only up to a certain \textit{inflection point}. Beyond this point (e.g., 12 bits in this specific configuration), further increases in precision yield negligible performance gains while continuing to increase hardware complexity. Allocating this high precision uniformly across all layers is therefore highly inefficient. This observation forms the basis for our layer-wise precision search.

\begin{figure*}[t]
    \centering
    \includegraphics[width=0.9\linewidth]{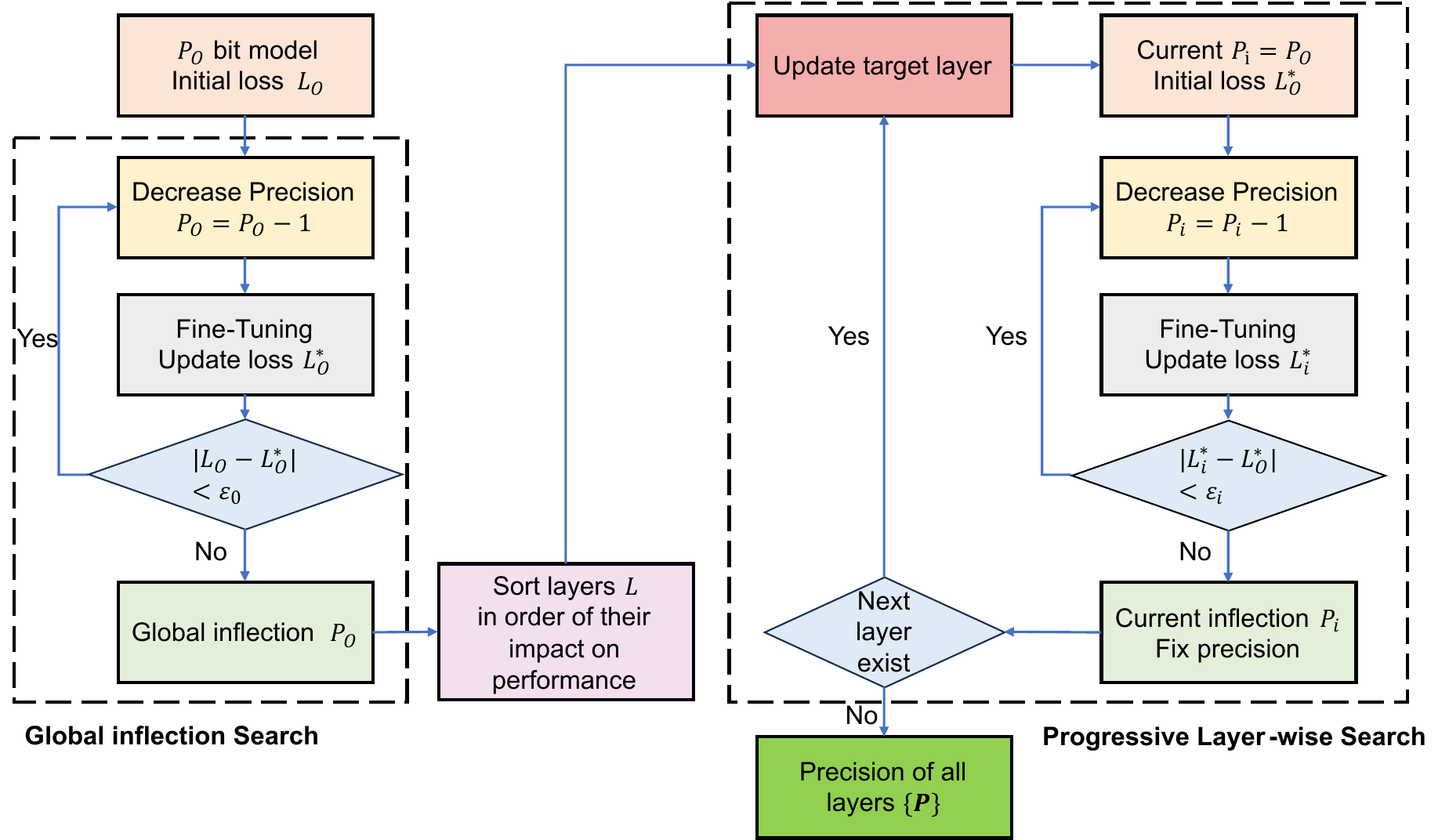}
    \caption{Flowchart of the Progressive Mixed-Precision Search algorithm. The process begins by establishing a high-precision baseline for the full network, then iteratively searches for the optimal, minimal bit-width for each layer without violating a set performance degradation tolerance $\varepsilon$.}
    \label{fig:search_flowchart}
\end{figure*}

We propose a Progressive Mixed-Precision Search algorithm to automatically determine the optimal bit-width for each layer in a structured and efficient manner. The algorithm, illustrated in Fig.~\ref{fig:search_flowchart}, proceeds as follows:

\begin{enumerate}
    \item Establish Performance Baseline: We first determine the inflection point for the entire network. We start with a high-precision model (e.g., 16-bit), which we assume has performance equivalent to the FP32 model, and record its converged loss. We then uniformly decrease the network's bit-width, fine-tuning at each step, until the performance degradation exceeds a predefined tolerance, $\varepsilon$. The last precision level that met the tolerance is set as the initial, high-precision baseline for all layers.
    \item 
    Progressive Layer-wise Search: Starting from the baseline, we iterate through the network layers in a predefined order (in order of their impact on performance). For each target layer:
    \begin{enumerate}
        \item We progressively decrease its bit-width while keeping all other layers at their current precision.
        \item After each decrease, we fine-tune the model for a few epochs.
        \item We find the lowest bit-width for the layer that does not violate the performance tolerance $\varepsilon_i$.
        \item This optimal bit-width is then fixed, and we proceed to the next layer.
    \end{enumerate}
    \item  Controlling the Trade-off: The tolerance parameter $\varepsilon$ provides direct control over the final performance-complexity trade-off. A smaller $\varepsilon$ yields a high-performance model with higher average bit-width, while a larger $\varepsilon$ produces a more compact model at the cost of some performance.
\end{enumerate}

This algorithm can be seamlessly integrated with our DRAQ framework by applying the weight regularization during all fine-tuning steps and re-calibrating the activation clipping parameters as needed.

\subsubsection{Equivalent Bit-width for Complexity Budgeting}
To make a fair comparison between a uniform-precision model and our mixed-precision models, we define an Equivalent Bit-width ($P_m$) metric. This metric computes the weighted average bit-width of a model, with weights determined by the memory footprint of each layer's feature map. For the synthesis transform, for instance, this is calculated as:

\begin{equation}
P_m = \frac{8}{15}P_1 + \frac{4}{15}P_2 + \frac{2}{15}P_3 + \frac{1}{15}P_4,
\end{equation}
where $P_i$ is the bit-width of the $i$-th layer. This metric allows us to create a mixed-precision model with, for example, an equivalent bit-width of 8 bits ($P_m \approx 8$), and compare its RD performance directly against a standard, uniform 8-bit model to demonstrate the benefits of our approach.

\subsection{Channel Pruning via GDN-Slimming}

While increasing model capacity by adding more channels is a standard method for improving the rate-distortion performance of LIC models, this approach leads to a direct and significant rise in computational and memory complexity. The choice of channel count presents a fundamental trade-off: larger models achieve better compression, particularly at high bit-rates, but their deployment on resource-constrained hardware becomes infeasible. This necessitates a more sophisticated approach to model optimization than simply scaling the architecture up or down.

Our motivation for channel pruning stems from a key observation: a large portion of this computational capacity is often redundant. A deeper analysis of the network's behavior reveals that across diverse input image sequences, a consistent subset of channels exhibits perpetually low activation energy. These channels contribute minimally to the model's final output, regardless of the input data. This consistent inactivity represents a significant opportunity for optimization. By systematically identifying and removing these redundant channels, we can create a much smaller, more efficient model that preserves the RD performance of the original, oversized architecture. This targeted pruning allows us to break the rigid trade-off between model size and performance, making SOTA compression models practical for hardware deployment.

To remove these redundant channels, we adapt the network \textit{slimming} methodology. The original slimming technique imposes L1 regularization on the scaling factors ($\gamma$) of Batch Normalization (BN) layers. During training, the $\gamma$ factors of unimportant channels are driven to zero, allowing them to be safely pruned. A central challenge, however, is that conventional channel slimming relies on the scaling factors within Batch Normalization (BN) layers, which are absent in LIC architectures.

To solve this, we modify the GDN layer to make it \textit{slimmable}. We introduce a learnable, channel-wise affine transformation immediately after the standard GDN operation, consisting of a scaling factor $a_i$ and a bias $b_i$ for each channel $i$:
\begin{align}
z_i^* &= a_i z_i+b_i \\
&= \frac{a_i x_i}{\beta_i + \sum_{j} \gamma_{ij} |x_j|}+b_i,
\end{align}

Here, the scaling factor $a_i$ serves the same purpose as $\gamma$ in a BN layer. We can now apply L1 regularization to the vector of $a_i$ factors during training:
\begin{equation}
L=R+\lambda D+\eta ||\textbf{a}||_1,
\end{equation}
where $\eta$ is the regularization strength. This process drives the scaling factors of unimportant channels to zero. After training, we prune all channels whose $a_i$ is below a small threshold $\varepsilon$, along with the corresponding filters in the preceding and succeeding convolutional layers. The resulting smaller network is then fine-tuned to recover any lost performance.

This entire optimization process is performed in sequence: we first apply GDN-slimming to get a structurally smaller model, and then apply the progressive mixed-precision search to the pruned model.

\section{Experiments}

\subsection{Experimental Setup}

\subsubsection{Datasets}

Models were trained on a 72,000-image subset of the PASS dataset \cite{asano2021pass} and tested on the standard Kodak24 dataset \cite{kodak24}.

\subsubsection{Evaluation Metrics}

We report performance using standard PSNR vs. bits-per-pixel (bpp) curves. We use the Bjøntegaard Delta Rate (BD-rate) to provide a single metric for the average bitrate difference (overhead) compared to a reference model. We also measure computational complexity in GFLOPs.

\subsubsection{Training Details}
All models were trained on a single NVIDIA RTX 3090 GPU. We trained on random $256 \times 256$ image crops with a batch size of 8. We use the Adam optimizer with an initial learning rate of $1 \times 10^{-4}$ and default parameters ($\beta_1=0.9$, $\beta_2=0.999$). To evaluate rate-distortion performance across a range of bitrates, we train models optimized for MSE with five $\lambda$ values: $\{0.0018, 0.0035, 0.0067, 0.013, 0.025\}$.

For the three lower $\lambda$ values, the model hyperparameters were set to $N=128, M=192$, and models were trained for 1,000,000 iterations. For the two higher $\lambda$ values, we used a larger model with $N=192, M=320$ and trained for 2,000,000 iterations. The fine-tuning stage for our dynamic range-aware methods consisted of an additional 500,000 iterations.

\subsubsection{Baselines and Compared Methods}
To validate our approach, we conduct a comprehensive set of experiments.

\begin{itemize}
    \item \textbf{Backbone Architectures:} We evaluate our quantization methods on two LIC architectures:
    \begin{itemize}
        \item The seminal hyperprior model from Ballé et al.~\cite{balle2018variational}.
        \item The checkerboard context model from He et al.~\cite{he2021checkerboard}.
    \end{itemize}

    \item \textbf{Comparison Framework:} To demonstrate the effectiveness of our proposed methods, we establish the following comparisons:
    \begin{itemize}
        \item \textbf{A Baseline Quantization Method:} We apply the standard Quantization-Aware Training (QAT) scheme from~\cite{jacob2018quantization} without any of our proposed distribution control techniques. This serves to isolate the performance gains attributable to our method.
        
        \item \textbf{An Ablation Study on GDN vs. ReLU:} To highlight the importance of our GDN-specific optimizations, we compare the performance of both ReLU-based and GDN-based backbones after applying our quantization framework.
        
        \item \textbf{A Comparison with former Methods:} We compare our final models against other published LIC quantization methods, specifically those from Hong et al.~\cite{hong2020efficient} and Sun et al.~\cite{sun2021end}. To ensure a fair comparison, all methods are applied to the same ReLU-based backbone, consistent with their original papers.
    \end{itemize}
\end{itemize}

\subsection{Foundational Quantization Performance (DRAQ)}
We evaluate our proposed Dynamic Range-Aware Quantization (DRAQ) framework on two SOTA backbones, Ballé2018~\cite{balle2018variational} and He2021~\cite{he2021checkerboard}. We analyze its effectiveness against a baseline quantization method and its unique ability to unlock the performance of GDN-based models.

\subsubsection{Performance vs. Baseline and Full-Precision}

\begin{figure*}
    \centering
    \subfloat[Ballé2018: Quantized vs. FP32]{
        \label{fig:rd_balle}
        \includegraphics[width=0.35\linewidth]{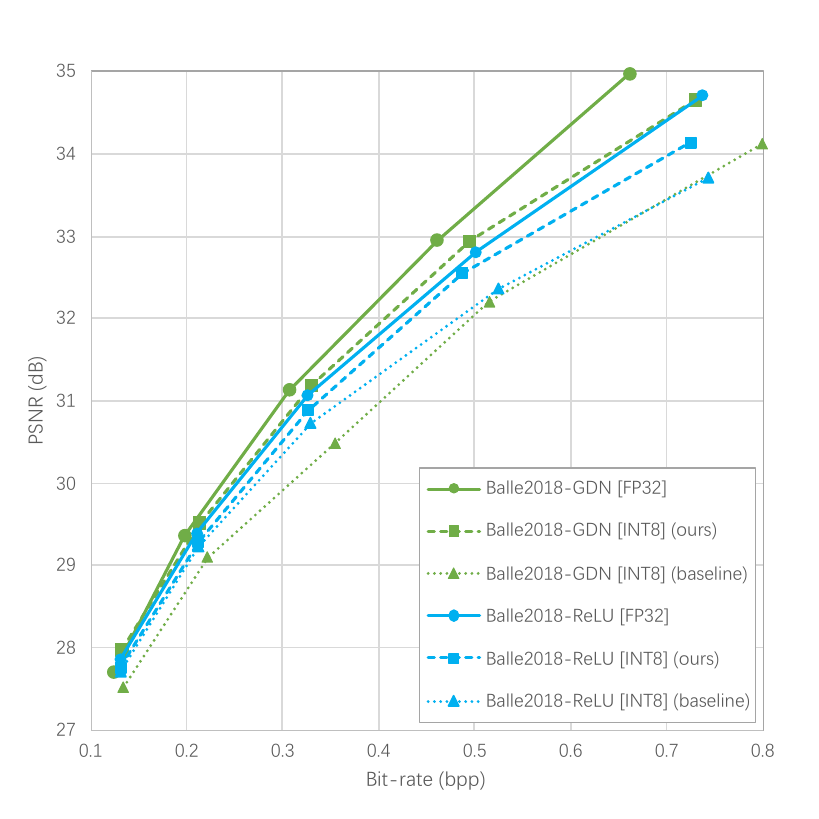}
    }
    \hspace{2cm}
    \subfloat[He2021: Quantized vs. FP32]{
        \label{fig:rd_he}
        \includegraphics[width=0.35\linewidth]{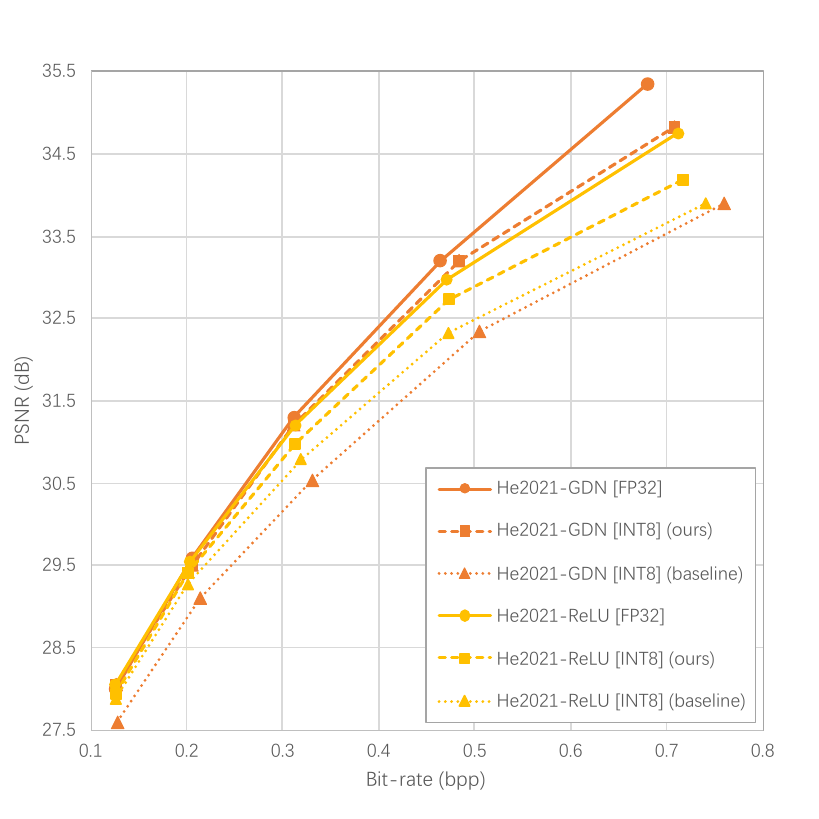}
    }
    \caption{Rate-distortion performance comparison. Our proposed DRAQ method significantly outperforms the baseline quantization and nearly matches the performance of the full-precision models for both ReLU and GDN-based architectures.}
    \label{fig:rd_curves}
\end{figure*}

Fig~\ref{fig:rd_curves} presents the core rate-distortion (RD) performance of our method. The results clearly demonstrate that for both the Ballé2018 and He2021 architectures, our DRAQ framework  significantly outperforms the baseline quantization scheme. Our method substantially closes the performance gap to the full-precision (FP32) models, especially at lower bitrates, validating the effectiveness of our distribution control techniques.

To quantify these improvements, we report the Bjøntegaard Delta Rate (BD-rate) in Table~\ref{tab:bd_rate_results}. BD-rate measures the average percentage of bitrate savings for a given PSNR compared to a reference model (here, the FP32 model); a smaller value indicates better performance. Our proposed method achieves a remarkable reduction in BD-rate across all tested configurations. For instance, on the Ballé2018-ReLU model, DRAQ reduces the BD-rate from the baseline's 12.8\% to just 4.93\%. Similarly, for the more complex He2021-GDN model, the performance gap is reduced from a prohibitive 24.0\% to a much more manageable 9.48\%.

\begin{table}[h]
\centering
\caption{BD-Rate (\%) vs. Full-Precision Models. A smaller value indicates better performance. Our proposed method consistently and significantly reduces the performance gap.}
\label{tab:bd_rate_results}
\begin{tabular}{llcc}
\toprule
\textbf{Model} & \textbf{Quantization Method} & \textbf{Precision} & \textbf{BD-Rate (\%)} \\
\midrule
\multirow{4}{*}{Ballé2018-ReLU}
& Hong2020~\cite{hong2020efficient} & INT10 & 26.5 \\
& Baseline & INT8 & 12.8 \\ 
& Sun2021~\cite{sun2021end} & INT8 & 9.01 \\
& \textbf{Proposed (DRAQ)} & \textbf{INT8} & \textbf{4.93} \\ 
\midrule
\multirow{2}{*}{Ballé2018-GDN}   
& Baseline & INT8 & 30.0 \\ 
& \textbf{Proposed (DRAQ)} & \textbf{INT8} & \textbf{6.30} \\ 
\midrule
\multirow{2}{*}{He2021-ReLU}     
& Baseline & INT8 & 15.3 \\ 
& \textbf{Proposed (DRAQ)} & \textbf{INT8} & \textbf{6.82} \\ 
\midrule
\multirow{2}{*}{He2021-GDN}     
& Baseline & INT8 & 24.0 \\ 
& \textbf{Proposed (DRAQ)} & \textbf{INT8} & \textbf{9.48} \\ 
\bottomrule
\end{tabular}
\end{table}

\subsubsection{Unlocking GDN Performance in the Quantized Domain}

A key achievement of our work is enabling high-performance quantization for GDN-based models, which typically suffer greatly from standard quantization. As seen in Fig.~\ref{fig:rd_curves}, with the baseline method, the quantized GDN models (e.g., dashed green line) perform \textit{worse} than their simpler ReLU counterparts (e.g., dashed blue line). 

However, when our DRAQ framework is applied, this trend is reversed. The quantized GDN models consistently outperform their quantized ReLU counterparts, mirroring the behavior of the FP32 models. This demonstrates that our specialized GDN quantization and clipping strategy is effective. Table~\ref{tab:gdn_vs_relu_points} provides a point-by-point comparison, confirming that at similar bitrates, our quantized GDN models achieve superior PSNR across all tested conditions.

\begin{table}[h]
\centering
\caption{Performance Comparison of Quantized ReLU and GDN Models Using Our Proposed DRAQ Framework.}
\label{tab:gdn_vs_relu_points}
\begin{tabular}{cccccc}
\toprule
& & \multicolumn{2}{c}{\textbf{Quantized ReLU}} & \multicolumn{2}{c}{\textbf{Quantized GDN}} \\ 
\cmidrule(lr){3-4} \cmidrule(lr){5-6}
\textbf{Model} & $\boldsymbol{\lambda}$ & \textbf{BPP} & \textbf{PSNR (dB)} & \textbf{BPP} & \textbf{PSNR (dB)} \\ 
\midrule
\multirow{4}{*}{Ballé2018}
& 0.0018 & 0.13 & 27.75 & 0.13 & \textbf{27.97} \\
& 0.0035 & 0.20 & 29.28 & 0.21 & \textbf{29.51} \\
& 0.0067 & 0.31 & 30.88 & 0.33 & \textbf{31.18} \\
& 0.0130 & 0.49 & 32.55 & 0.50 & \textbf{32.94} \\ 
\midrule
\multirow{4}{*}{He2021}
& 0.0018 & 0.13 & 27.93 & 0.13 & \textbf{28.05} \\
& 0.0035 & 0.20 & 29.40 & 0.20 & \textbf{29.52} \\
& 0.0067 & 0.31 & 30.97 & 0.31 & \textbf{31.21} \\
& 0.0130 & 0.47 & 32.73 & 0.48 & \textbf{33.20} \\ 
\bottomrule
\end{tabular}
\end{table}

\subsubsection{Comparison with Existing Methods}
Finally, we benchmark our method against other published LIC quantization techniques on the Ballé2018-ReLU backbone. As shown in Table~\ref{tab:bd_rate_results}, our method achieves a BD-rate of just 4.93\%, substantially outperforming both Hong et al.~\cite{hong2020efficient} (26.5\%) and Sun et al.~\cite{sun2021end} (9.01\%). Notably, our method achieves this superior performance at INT8 precision, whereas the method of Hong et al. required INT10, further highlighting the efficiency of our approach.

\subsubsection{Ablation Study}
To validate the individual contributions of our proposed components—activation range clipping and weight outlier regularization—we conduct a thorough ablation study. We compare the performance of our full DRAQ framework against three variants: 
\begin{enumerate}
    \item A \textbf{Baseline} model with standard QAT.
    \item A model with only \textbf{Weight Regularization}.
    \item A model with only \textbf{Activation Clipping}.
\end{enumerate}

\begin{table}[h]
\centering
\caption{Ablation Study Results (BD-Rate \% vs. FP32). Lower is better. The results show that both components contribute to performance, with their combination being the most effective. The minimal gain from clipping in the FP32 domain confirms the improvement is primarily due to better quantization.}
\label{tab:ablation_results}
\begin{tabular}{llcc}
\toprule
\textbf{Model} & \textbf{Ablation Condition} & \textbf{Precision} & \textbf{BD-Rate (\%)} \\ 
\midrule
\multirow{4}{*}{Ballé2018-ReLU}
& Baseline (No Clipping/Reg) & INT8 & 12.8 \\
& Weight Reg Only & INT8 & 10.8 \\
& Activation Clip Only & INT8 & 6.16 \\ 
& \textbf{Proposed (Both)} & \textbf{INT8} & \textbf{4.93} \\ 
\midrule
\multirow{6}{*}{He2021-GDN}     
& Baseline (No Clipping/Reg) & INT8 & 24.0 \\
& Weight Reg Only & INT8 & 18.9 \\
& Activation Clip Only & INT8 & 11.3 \\ 
& \textbf{Proposed (Both)} & \textbf{INT8} & \textbf{9.48} \\ 
\cmidrule{2-4}
& FP32 (No Clipping) & FP32 & (Reference) \\ 
& FP32 (With Clipping) & FP32 & -1.02 \\ 
\bottomrule
\end{tabular}
\end{table}

\begin{figure}[t]
    \centering
    \includegraphics[width=0.95\linewidth]{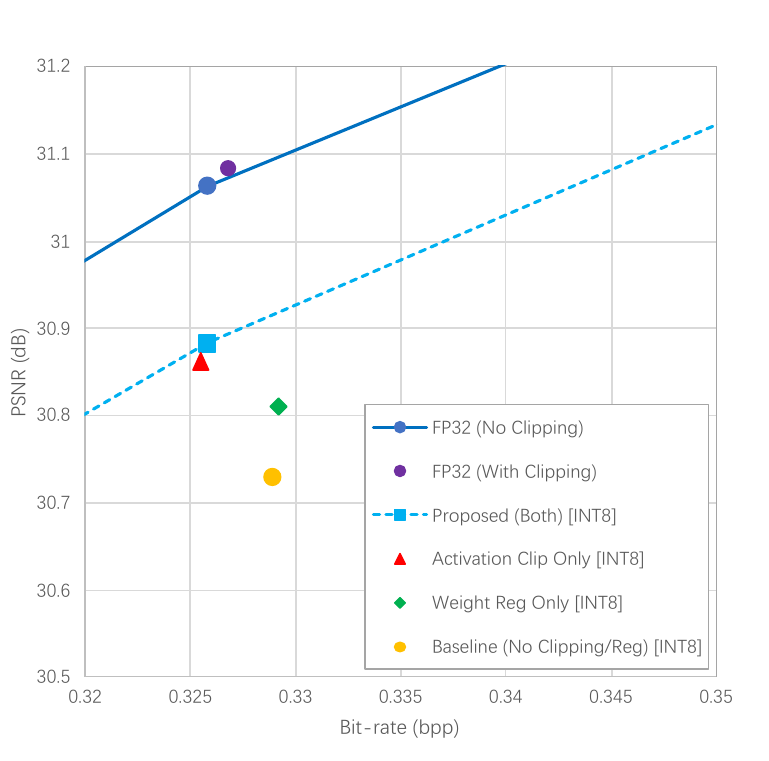}
    \caption{Visual results of the ablation study on the Ballé2018-ReLU model. The full DRAQ framework (both components, purple triangles) achieves the best rate-distortion trade-off, clearly outperforming the baseline and either individual component.}
    \label{fig:ablation_curves}
\end{figure}

The results, summarized in Table~\ref{tab:ablation_results}, clearly demonstrate the effectiveness of both components. For the Ballé2018-ReLU model, starting from a baseline BD-rate of 12.8\%, adding only weight regularization reduces it to 10.8\%, while activation clipping alone achieves a more significant reduction to 6.16\%. However, combining both components in our full DRAQ framework yields the best performance, achieving a BD-rate of just 4.93\%. This synergy confirms that addressing both weight and activation distributions is necessary for optimal performance. Figure~\ref{fig:ablation_curves} visually corroborates these findings.

Furthermore, a crucial aspect of this study is to determine if the performance gains stem from creating a more quantization-friendly distribution or simply from an architectural improvement (i.e., that adding a \textit{Clip} layer inherently improves the FP32 model). To investigate this, we applied our clipping strategy directly to the full-precision He2021-GDN model. As shown in the final rows of Table~\ref{tab:ablation_results}, adding clipping to the FP32 model yields only a minor BD-rate improvement of -1.02\%. In stark contrast, applying the same clipping strategy in the quantized domain reduces the BD-rate penalty from 24.0\% down to 11.3\%—a massive improvement. This disparity confirms that the vast majority of the performance gain is indeed due to creating a more efficient data distribution for quantization, not merely from enhancing the FP32 model's architecture.

\subsubsection{Analysis of Quantization Error and Data Distribution}
Having established the performance benefits of our DRAQ framework, we now provide a direct analysis to verify its underlying mechanism: that it improves performance by fundamentally reshaping the data distributions to be more quantization-friendly.

\textbf{Reduction of Quantization Error:}
We first quantitatively measure the impact of our method on the mean squared quantization error (MSQE). Table~\ref{tab:quant_error_analysis} compares the MSQE for two representative convolutional layers before (A, B) and after (A*, B*) applying our DRAQ framework. The results show a marked reduction in quantization error for both weights and activations. For instance, in Layer B, our method reduces the activation MSQE by over 60\%, from $6.46e-6$ to $2.48e-6$. This directly confirms that our distribution control techniques lead to a more accurate low-bit representation.

\begin{table}[h]
\centering
\caption{Mean Squared Quantization Error (MSQE) for two representative layers before (A, B) and after (A*, B*) applying our DRAQ framework. Our method significantly reduces the error for both weights and activations.}
\label{tab:quant_error_analysis}
\begin{tabular}{lcccc}
\toprule
 & {A} & {A*} & {B} & {B*} \\
\midrule
\textbf{Weight}     & 2.10e-6 &  2.02e-6 & 7.55e-7 &  6.36e-7 \\
\textbf{Activation} & 2.07e-6 &  1.03e-6 & 6.46e-6 &  2.48e-6 \\
\bottomrule
\end{tabular}
\end{table}

\begin{figure*}
    \centering
    \subfloat[Weight distribution before and after applying regularization.]{
    \label{fig:dist_weights}
    \includegraphics[width=0.4\linewidth]{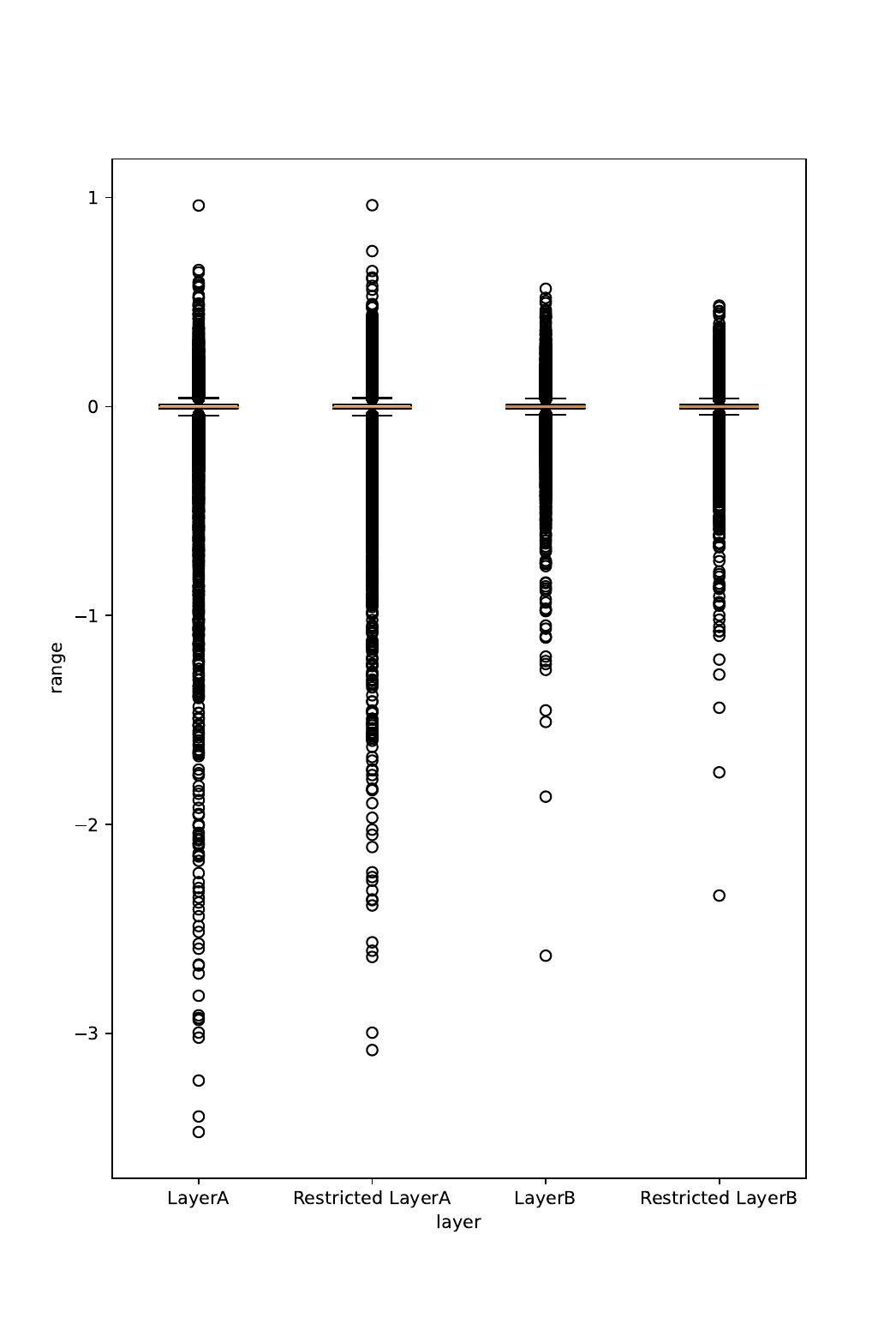}
    }
    \subfloat[Activation distribution before and after applying clipping.]{
    \label{fig:dist_activations}
    \includegraphics[width=0.4\linewidth]{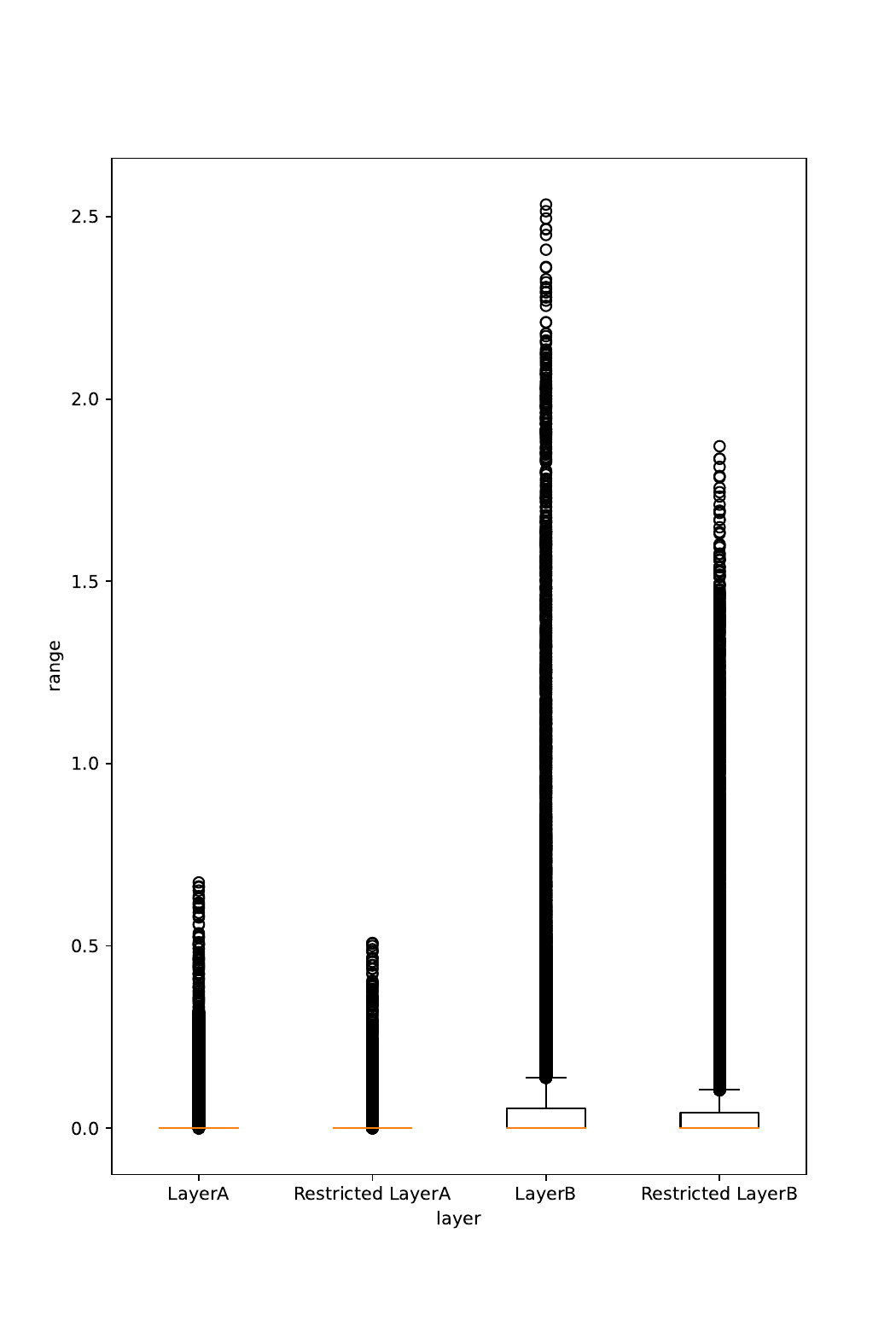}
    }
    \caption{Visual analysis of data distribution reshaping. Our proposed methods successfully (a) reduce the range and outliers of weights and (b) constrain the dynamic range of activations, making them more amenable to quantization.}
    \label{fig:distribution_analysis}
\end{figure*}

\textbf{Verification of Distribution Reshaping:}
Figure~\ref{fig:distribution_analysis} provides a visual confirmation of how this error reduction is achieved. The boxplots illustrate the distributions of weights and activations for a convolutional layer before and after applying DRAQ. As hypothesized, our method effectively reshapes the distributions:
\begin{itemize}
    \item \textbf{For weights (Fig.~\ref{fig:dist_weights})}, the regularization dramatically compresses the value range by pruning the most extreme outliers, as seen by the reduced span of the whiskers and fewer outlier points.
    \item \textbf{For activations (Fig.~\ref{fig:dist_activations})}, the calibrated clipping successfully tames the overall dynamic range, pulling the maximum values down significantly.
\end{itemize}
These visualizations directly correspond to our design goals. By creating narrower and more concentrated data distributions, our DRAQ framework allows the fixed number of quantization bins to represent the data with much higher fidelity, which explains the superior rate-distortion performance observed in our main results.

\subsection{Mixed-Precision Quantization Results}

We now evaluate the performance of our Progressive Mixed-Precision Search algorithm. We demonstrate its effectiveness in two key scenarios: achieving near-lossless performance relative to the FP32 model, and optimizing performance under a fixed complexity budget.

\subsubsection{Achieving Near-Lossless Performance}
First, we demonstrate the algorithm's ability to find a bit-width configuration that minimizes the performance gap to the full-precision model. By setting a strict performance tolerance ($\varepsilon$), the search identifies the optimal precision for each layer.

\begin{figure}
    \centering
    \subfloat[Optimal bit-widths for the baseline Ballé2018 model ($\lambda=0.0067$).]{
    \label{fig:mix_baseline}
    \includegraphics[width=0.85\linewidth]{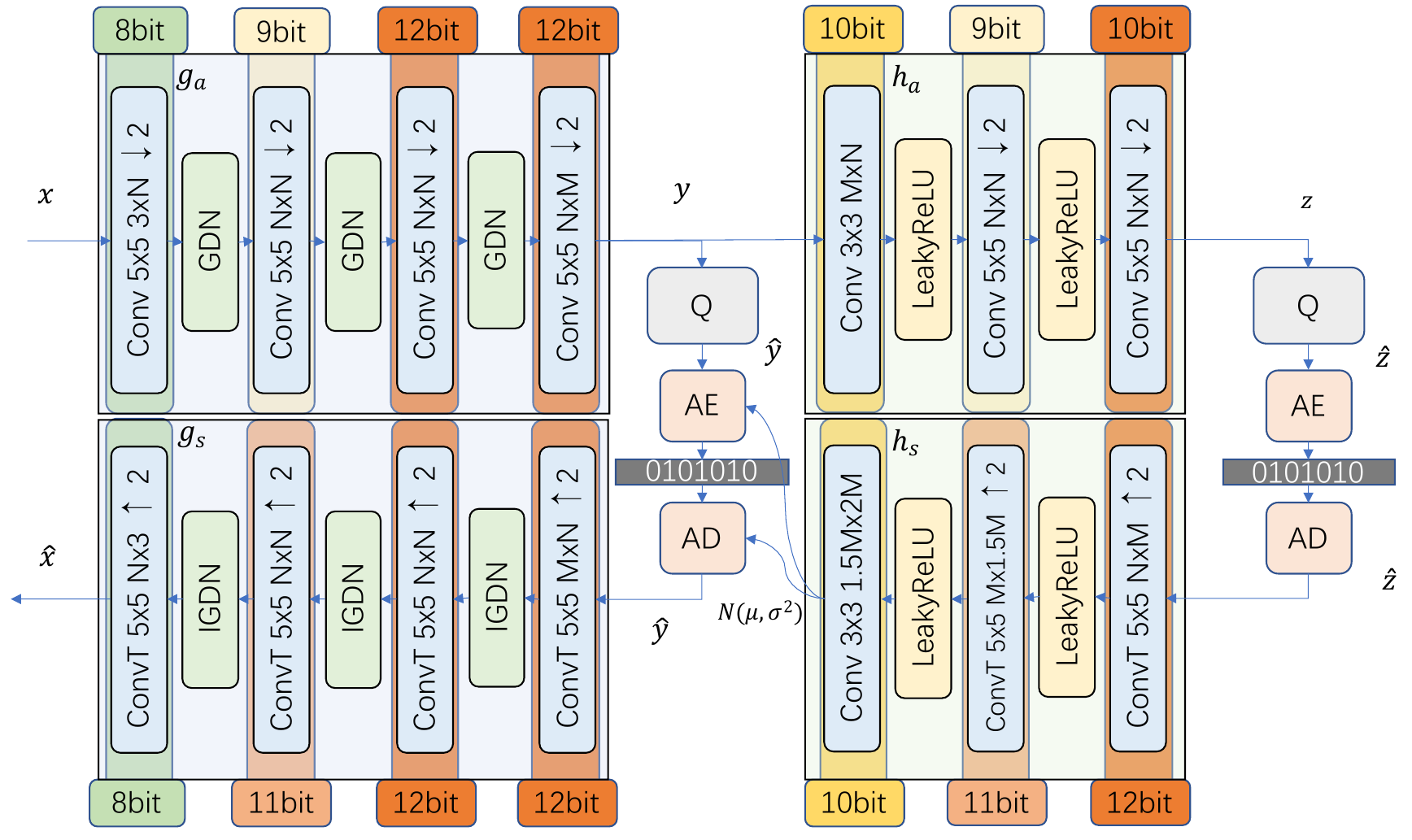}
    }
    \\
    \subfloat[Optimal bit-widths for the DRAQ-enhanced Ballé2018 model ($\lambda=0.025$).]{
    \label{fig:mix_draq}
    \includegraphics[width=0.85\linewidth]{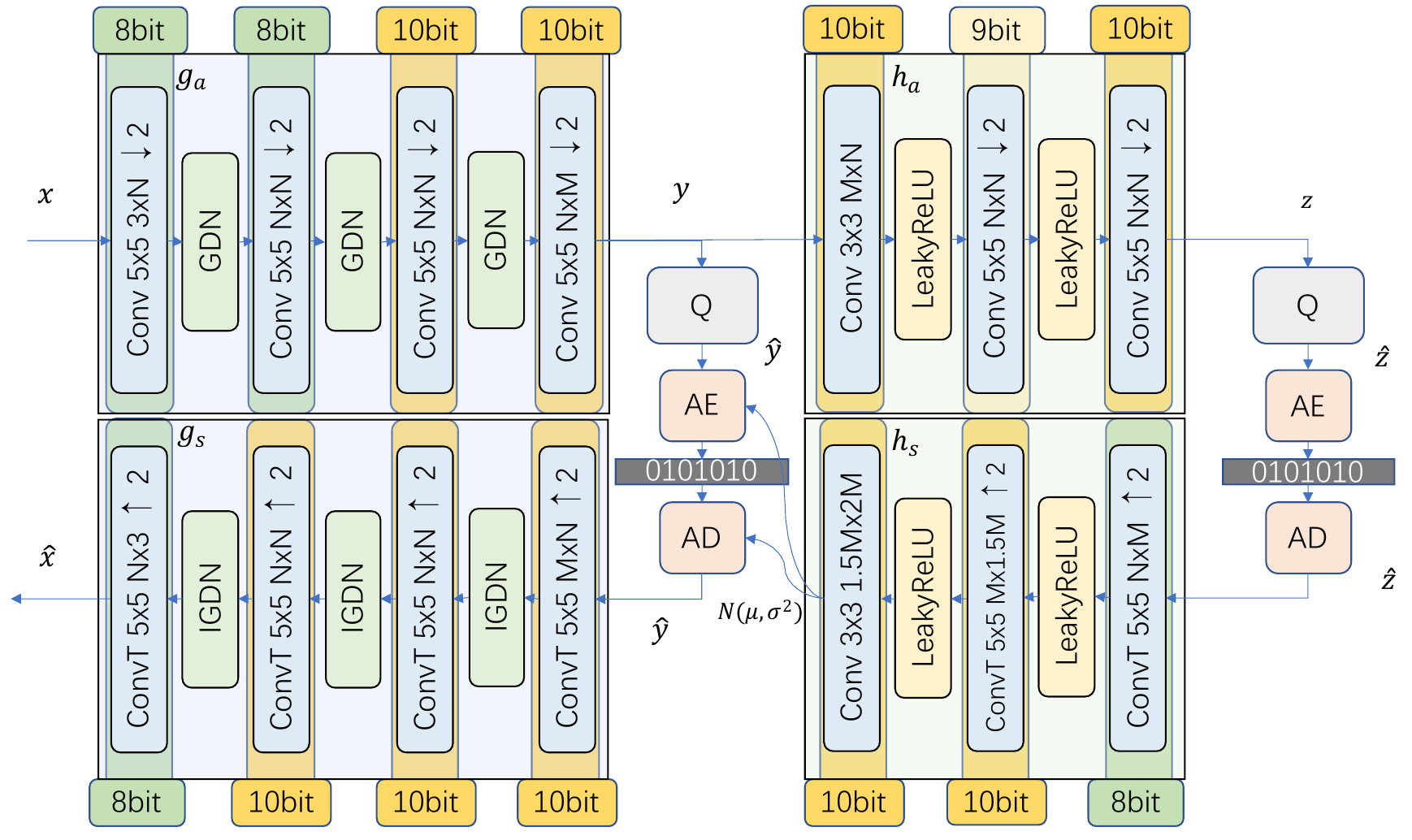}
    }
    \caption{Mixed-precision assignments that achieve near-lossless performance. Applying our DRAQ framework first (b) significantly reduces the required bit-widths compared to the baseline (a) to achieve performance comparable to the FP32 model.}
    \label{fig:mixed_precision_configs}
\end{figure}

Figure~\ref{fig:mixed_precision_configs} illustrates the resulting bit-width assignments for the Ballé2018 model. A key finding emerges when comparing the results with and without our DRAQ framework. For a baseline quantized model (Fig.~\ref{fig:mix_baseline}), the optimal bit-widths are primarily between 10 and 12 bits. However, when our DRAQ framework is first applied (Fig.~\ref{fig:mix_draq}), the required precision for near-lossless performance drops significantly to a range of 8 to 10 bits. This synergy demonstrates that our DRAQ method not only improves uniform-precision models but also makes them more amenable to further compression via mixed-precision quantization.

The performance for the configuration in Fig.~\ref{fig:mix_draq} is shown in the second row of Table~\ref{tab:mixed_precision_summary}. With an equivalent bit-width of only 8.7 bits, the model achieves a PSNR of 34.62 dB, negligibly close to the FP32 performance of 34.71 dB.

\subsubsection{Optimizing for a Fixed Complexity Budget}
\begin{figure}[h]
    \centering
    \includegraphics[width=0.9\linewidth]{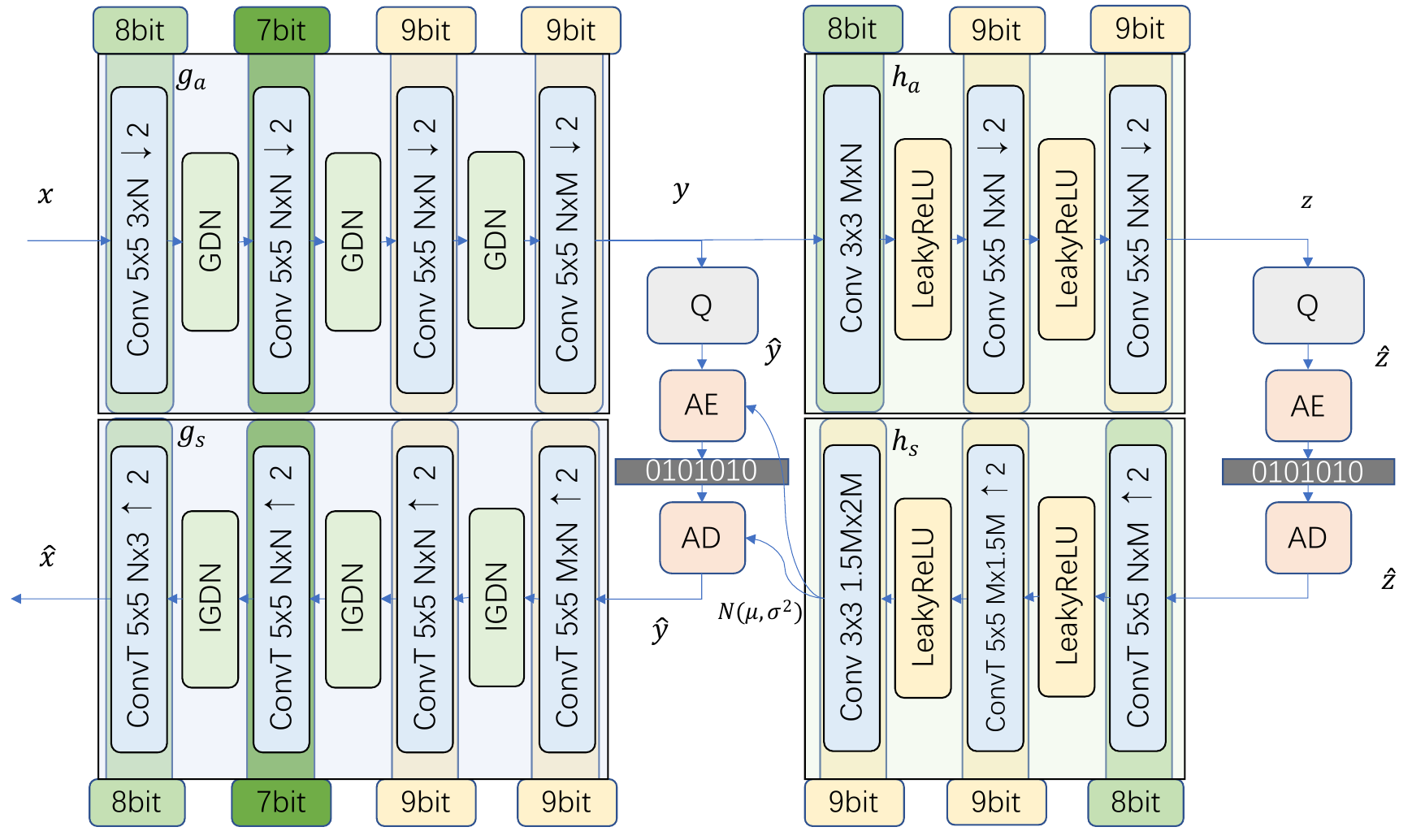}
    \caption{Mixed-precision assignment for the Ballé2018 model targeting an equivalent bit-width of approximately 8 bits.}
    \label{fig:mixed_8bit_equivalent}
\end{figure}

A key advantage of our method is its ability to optimize performance for a specific hardware budget. To demonstrate this, we relaxed the tolerance $\varepsilon$ and targeted an equivalent bit-width of approximately 8 bits. The resulting layer-wise precision assignment is shown in Fig.~\ref{fig:mixed_8bit_equivalent}.

Table~\ref{tab:mixed_precision_summary} compares the performance of this configuration against the uniform 8-bit model. At a nearly identical bitrate and complexity budget, our mixed-precision model ($P_m=7.9$ bits) achieves a PSNR of 34.35 dB, which is a significant improvement over the 34.13 dB from the uniform 8-bit model. This result highlights the primary benefit of mixed-precision quantization for hardware deployment: for a given complexity budget, intelligently reallocating bits to more sensitive layers yields a superior rate-distortion trade-off.

\begin{table}[h]
\centering
\caption{Performance Comparison of Mixed-Precision Configurations Against FP32 and Uniform 8-bit Models.}
\label{tab:mixed_precision_summary}
\begin{tabular}{lccc}
\toprule
\textbf{Configuration} & \textbf{(Equiv.) Bit-Width} & \textbf{BPP} & \textbf{PSNR (dB)} \\
\midrule
Full Precision (FP32) & 32 & 0.737 & 34.71 \\
Mixed Precision (Near-Lossless) & 8.7 & 0.725 & 34.62 \\
Mixed Precision (Budgeted) & 7.9 & 0.726 & 34.35 \\
Uniform Quantization & 8.0 & 0.725 & 34.13 \\
\bottomrule
\end{tabular}
\end{table}

\subsection{GDN-Slimming Results}
We evaluate our GDN-Slimming technique to validate its ability to reduce model complexity while preserving rate-distortion (RD) performance.

\subsubsection{Analysis of Redundancy Removal}

First, we verify that the method correctly identifies and removes redundant channels based on model capacity and task difficulty. We trained several He2021 models with different initial channel counts and at different RD trade-off points ($\lambda$).

As shown in Table~\ref{tab:pruning_channels}, our method demonstrates intelligent, adaptive pruning. When two models with different initial sizes ($N=M = 192$ vs. $N=128, M=192$) are pruned for the same low-bitrate task, the resulting channel counts become much more similar, indicating that the method effectively removes initial over-parameterization. 

\begin{table}[h]
\centering
\caption{Channel counts of the three main GDN layers before and after slimming, demonstrating adaptive redundancy removal.}
\label{tab:pruning_channels}
\begin{tabular}{ccc}
\toprule
$\boldsymbol{\lambda}$ & \textbf{Initial Channels} & \textbf{Pruned Channels} \\
\midrule
\multirow{2}{*}{0.0035} & [192, 192, 192] & [110, 144, 182] \\
& [128, 128, 128] & [114, 121, 125] \\
\midrule
0.0067 & [192, 192, 192] & [142, 173, 191] \\
0.0130 & [192, 192, 192] & [172, 188, 190] \\
\bottomrule
\end{tabular}
\end{table}

\begin{figure}[t]
    \centering
    \includegraphics[width=0.5\linewidth]{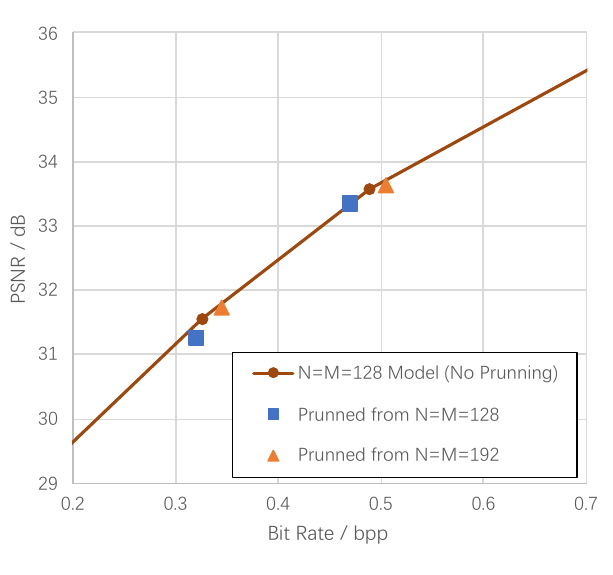}
    \caption{Rate-distortion performance of pruned models. Models with different initial sizes (e.g., N=192 vs. N=128) converge to a nearly identical RD curve after pruning. This demonstrates that the slimming method effectively removes structural redundancy without compromising performance.}
    \label{fig:slim_rd_curves}
\end{figure}

Critically, this reduction in complexity comes at a negligible cost to RD performance. Figure~\ref{fig:slim_rd_curves} shows that the pruned models achieve nearly identical RD curves compared to their unpruned counterparts.

\subsubsection{Performance vs. Complexity Trade-off}

The primary benefit of pruning is achieving a more favorable balance between performance and computational cost. Table~\ref{tab:complexity_comparison} benchmarks our pruned GDN-Slimming model against other SOTA architectures. We measure complexity in GFLOPs per pixel and performance in terms of bitrate savings relative to the BPG codec.

The results are compelling. Compared to the original He2021 model, our GDN-Slimming model reduces the total computational complexity by approximately 20\% (from 138.5 to 109.9 GFLOPs/pixel). This significant reduction is achieved with only a minimal drop in bitrate savings (from 11.29\% to 10.66\%), and a slight reduction in encoding time. This positions our pruned model as an excellent and practical trade-off point, offering performance substantially better than Ballé2018 at a lower complexity than He2021.

\begin{table}[h]
\centering
\caption{Complexity and Performance Comparison with LIC Models. Our GDN-Slimming model offers a superior trade-off point.}
\label{tab:complexity_comparison}
\begin{tabular}{lcccc}
\toprule
\textbf{Model} & \textbf{Total GFLOPs/pixel} & \textbf{Bitrate Savings (\%)} & \textbf{Time (ms)} \\
\midrule
Minnen2018~\cite{minnen2018joint} & 138.5 & 13.53 & 2.300 \\
He2021~\cite{he2021checkerboard} & 138.5 & 11.29 & 0.072 \\
\textbf{GDN-Slimming (Ours)} & \textbf{109.9} & \textbf{10.66} & \textbf{0.067} \\
Ballé2018~\cite{balle2018variational} & 120.8 & 5.72  & 0.062 \\
\bottomrule
\end{tabular}
\end{table}

\subsection{Final Integrated System Performance}

In this section, we evaluate the synergistic performance of our complete framework, which integrates Dynamic Range-Aware Quantization (DRAQ), Progressive Mixed-Precision Search, and GDN-Slimming.

\subsection{Comparison with State-of-the-Art Implementations}

We first benchmark our fully integrated framework against other published FPGA-based LIC implementations. Table~\ref{tab:sota_hardware_comparison} compares our solution to F-LIC~\cite{koyuncu2022device} and FPX-NIC~\cite{jia2022fpx}. Despite targeting a similar BPP, our method achieves the highest PSNR (30.14 dB) while requiring significantly lower computational complexity. Our model's encoding pipeline demands only 75.62 GFLOPs, demonstrating superior hardware efficiency compared to its counterparts.

\begin{table}[h]
\centering
\caption{Comparison with other FPGA-based LIC implementations. Our proposed framework achieves the best PSNR with the lowest computational complexity.}
\label{tab:sota_hardware_comparison}
\begin{tabular}{lccc}
\toprule
 & \textbf{Proposed Framework} & F-LIC~\cite{koyuncu2022device} & FPX-NIC~\cite{jia2022fpx} \\
\midrule
\textbf{Hardware Platform} & (N/A) & VCU118 & ZCU104   \\
\textbf{Model Precision} & INT7.9 (Mixed) & INT8 & INT8  \\
\midrule
\textbf{PSNR (dB)} & \textbf{30.14} & 29.94 & 29.89  \\
\textbf{BPP} & 0.2574 & 0.2542 & 0.2565  \\
\midrule
\textbf{GFLOPs ($g_a$ only)} & \textbf{71.64} & 91.48 & 380.44   \\
\textbf{Total Encoding GFLOPs} & \textbf{75.62} & 96.51 & 393.53  \\
\bottomrule
\end{tabular}
\end{table}

\subsection{Performance vs. Complexity Trade-off}

To clearly illustrate the benefits of our hardware-aware optimizations (mixed-precision and pruning), Table~\ref{tab:framework_ablation} provides a detailed comparison between the full-precision model, our DRAQ quantization, and the final integrated framework.

While the baseline DRAQ method successfully quantizes the model to INT8 with only a minor performance drop, our final framework achieves a remarkable trade-off. For example, at $\lambda=0.0130$, our mixed-precision and pruned model reduces the complexity of the main analysis transform ($g_a$) by over 50\% (from 312 to 149 GFLOPs) compared to the DRAQ model, while the PSNR decreases by only 0.31 dB. This demonstrates the efficacy of our hardware-aware optimizations in dramatically reducing computational load while maintaining performance close to the full-precision model, validating the core contribution of this work. By first creating a high-fidelity integer model and then applying targeted, hardware-aware optimizations, we have produced a final model that is verifiably superior in both efficiency and quality.

\begin{table*}[h]
\centering
\caption{Performance and complexity trade-offs. Our final framework significantly reduces computational complexity compared to both the FP32 and baseline quantized models (DRAQ) with only a minor impact on RD performance.}
\label{tab:framework_ablation}
\begin{tabular}{cccccc}
\toprule
\textbf{Method} & $\boldsymbol{\lambda}$ & \textbf{(Equiv.) Precision} & \textbf{$g_a$ GFLOPs} & \textbf{BPP} & \textbf{PSNR (dB)}  \\ 
\midrule
\multirow{4}{*}{Full Precision} & 0.0018 & \multirow{4}{*}{FP32} & 105.79 & 0.1253 & 28.05  \\
& 0.0035 & & 105.79 & 0.2033 & 29.54 \\
& 0.0067 & & 105.79 & 0.3134 & 31.20 \\
& 0.0130 & & 311.96 & 0.4707 & 32.97 \\ 
\midrule
\multirow{4}{*}{DRAQ} & 0.0018 & \multirow{4}{*}{INT8} & 105.79 & 0.1266 & 27.93  \\
& 0.0035 & & 105.79 & 0.2011 & 29.41 \\
& 0.0067 & & 105.79 & 0.3129 & 30.97 \\
& 0.0130 & & 311.96 & 0.4731 & 32.73  \\ 
\midrule
\multirow{4}{*}{\textbf{Proposed}} & 0.0018 & INT7.9 & \textbf{61.26} & 0.1250 & 27.81  \\
& 0.0035 & INT8.0 & \textbf{65.04} & 0.2029 & 29.28 \\
& 0.0067 & INT7.9 & \textbf{71.64} & 0.3138 & 31.03 \\
& 0.0130 & INT8.1 & \textbf{149.44} & 0.4743 & 32.42 \\ 
\bottomrule
\end{tabular}
\end{table*}

\section{Conclusion and Future Work}

Deploying high-performance learned image compression on resource-constrained hardware like FPGAs requires more than a simple application of quantization. In this paper, we have presented a comprehensive, multi-stage optimization framework that addresses this challenge from the ground up.

We began by identifying and solving the fundamental data distribution problems—extreme weight outliers and large activation dynamic ranges—that cause catastrophic performance loss in standard quantization. Our proposed Dynamic Range-Aware Quantization (DRAQ) framework successfully creates a high-fidelity 8-bit integer model that performs nearly identically to its full-precision floating-point parent, solving a critical bottleneck for all LIC models and, for the first time, enabling the effective quantization of advanced GDN-based architectures.

Building on this robust foundation, we introduced two hardware-aware optimization techniques to maximize efficiency on FPGAs. The progressive mixed-precision search algorithm leverages the custom datapath flexibility of FPGAs to assign optimal, non-uniform bit-widths to each layer, minimizing complexity. Our novel adaptation of channel slimming to GDN-based models effectively prunes network redundancy, further reducing the computational load.

The final, fully optimized model demonstrates state-of-the-art performance, achieving a better rate-distortion tradeoff at a lower computational cost than existing FPGA-based implementations. This work provides a complete and validated pathway for transforming complex, high-performance LIC models into efficient, hardware-friendly solutions, paving the way for their practical application in real-world, low-power visual data systems. While this paper has demonstrated excellent results in simulation, a hardware deployment would allow for the analysis of real-world metrics such as throughput, latency, power consumption, and on-chip resource utilization. Such a deployment would provide a definitive validation of the practical benefits of our hardware-software co-design methodology.

\bibliography{main.bib}

\end{document}